%% file: main.tex
\documentclass[10pt,twocolumn,letterpaper]{article}

\usepackage[pagenumbers]{cvpr}      % To produce the REVIEW version
\usepackage[utf8]{inputenc} % allow utf-8 input
\usepackage[T1]{fontenc}    % use 8-bit T1 fonts
\usepackage{url}            % simple URL typesetting
\usepackage{booktabs}       % professional-quality tables
\usepackage{amsfonts} 
\usepackage{comment}% blackboard math symbols
\usepackage{nicefrac}       % compact symbols for 1/2, etc.
\usepackage{microtype}      % microtypography
\usepackage{xcolor}         % colors
\usepackage{amsmath,bm}
\usepackage{float}
\usepackage{multirow, graphicx}
\usepackage{graphicx}
\usepackage{tabularx, makecell, multirow} 
\usepackage{wrapfig}

\usepackage{algorithm}
\usepackage{comment}

\usepackage{algpseudocode}

\usepackage{subcaption}    % for subfigures

\usepackage{amsmath,amssymb}

\usepackage[dvipsnames]{xcolor}
\usepackage{xcolor,colortbl}
\usepackage{multirow, makecell}
\definecolor{Gray}{gray}{0.50}
\newcolumntype{g}{>{\columncolor{Gray}}c}
\definecolor{ffe1da}{RGB}{255,225,218}
\definecolor{F7E0D5}{RGB}{247,224,213}
\definecolor{darkF7E0D5}{RGB}{209,154,128}
\colorlet{Light}{White!0!F7E0D5}

\colorlet{tabfirst}{Green!25}
\definecolor{tabthird}{rgb}{1, 0.85, 0.7}
\definecolor{tabsecond}{rgb}{1, 0.96, 0.7}

\definecolor{tabfirst}{rgb}{0.79, 0.92, 1.0} % Light Blue with High Saturation
\definecolor{tabsecond}{rgb}{1.0, 0.86, 0.86} % Light Red with High Saturation
\definecolor{tabthird}{rgb}{0.9, 0.8, 1.0} % Light Purple with High Saturation

\definecolor{cvprblue}{rgb}{0.21,0.49,0.74}
\definecolor{mypurple}{rgb}{0.4549,0.145,0.99}
\usepackage[pagebackref,breaklinks,colorlinks,allcolors=cvprblue]{hyperref}

\def\paperID{2716} % *** Enter the Paper ID here
\def\confName{CVPR}
\def\confYear{2025}
\title{\textcolor{mypurple}{MAC-Ego3D}: \textcolor{mypurple}{M}ulti-\textcolor{mypurple}{A}gent Gaussian \textcolor{mypurple}{C}onsensus for \\ Real-Time Collaborative  \textcolor{mypurple}{Ego}-Motion and Photorealistic \textcolor{mypurple}{3D} Reconstruction}
\author{
    $\text{Xiaohao Xu}^{1}\thanks{Equal contribution}$ \quad
    $\text{Feng Xue}^{1\ast}$\quad
    $\text{Shibo Zhao}^{2}$\quad
    $\text{Yike Pan}^{1}$\quad 
    $\text{Sebastian Scherer}^{2}$\quad
    $\text{Xiaonan Huang}^{1}$\\
    $^{1}$University of Michigan, Ann Arbor\quad $^{2}$Carnegie Mellon University\\
    }

\begin{document}
\maketitle
\input{sec/0_abstract}    
\input{sec/1_intro}

\input{sec/2_related_work}

\input{sec/3_method}

\input{sec/4_experiment}

\input{sec/5_conclusion}

\clearpage

\input{sec/X_suppl}


\end{document}

%% file: sec/0_abstract.tex
\begin{abstract}
Real-time multi-agent collaboration for ego-motion estimation and high-fidelity 3D reconstruction is vital for scalable spatial intelligence. However, traditional methods produce sparse, low-detail maps, while recent dense mapping approaches struggle with high latency.
To overcome these challenges, we present \textbf{\textit{MAC-Ego3D}}, a novel framework for real-time collaborative photorealistic 3D reconstruction via {\textit{Multi-Agent Gaussian Consensus}}. MAC-Ego3D enables agents to independently construct, align, and iteratively refine local maps using a unified Gaussian splat representation. Through \textit{Intra-Agent Gaussian Consensus}, it enforces spatial coherence among neighboring Gaussian splats within an agent. For global alignment, parallelized \textit{Inter-Agent Gaussian Consensus}, which asynchronously aligns and optimizes local maps by regularizing multi-agent Gaussian splats, seamlessly integrates them into a high-fidelity 3D model. Leveraging Gaussian primitives, MAC-Ego3D supports efficient RGB-D rendering, enabling rapid inter-agent Gaussian association and alignment.
MAC-Ego3D bridges local precision and global coherence, delivering higher efficiency, largely reducing localization error, and improving mapping fidelity. It establishes a new SOTA on synthetic and real-world benchmarks, achieving a \textbf{15$\times$ increase in inference speed}, \textbf{order-of-magnitude reductions in ego-motion estimation error for partial cases}, and \textbf{RGB PSNR gains of 4 to 10 dB}. Our code will be made publicly available at \url{https://github.com/Xiaohao-Xu/MAC-Ego3D}.
\end{abstract}

%% file: sec/1_intro.tex
\section{Introduction}
\label{sec:introduction}

\begin{figure}[t!]
\centering \setlength{\abovecaptionskip}{0.2cm}
\includegraphics[width=0.48\textwidth]{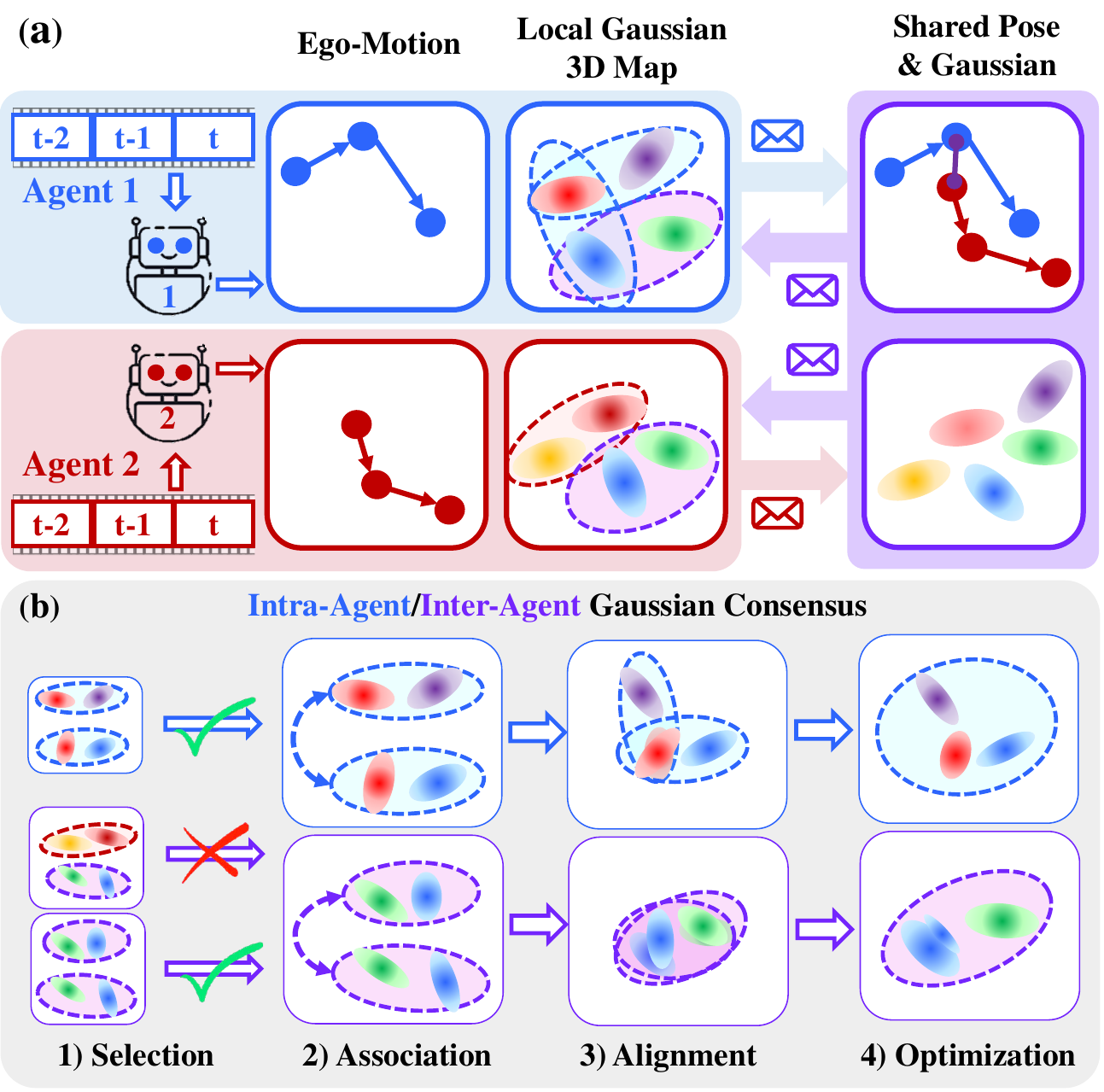}
\caption{\textcolor{black}{\textbf{ Towards collaborative, real-time, photorealistic 3D reconstruction in multi-agent systems.} (\textbf{a}) In \textbf{\textit{MAC-Ego3D}}, each agent independently captures observations, estimates ego-motion, and constructs a local Gaussian-based 3D map, which is then periodically aligned with others for collaborative optimization. (\textbf{b}) \textit{\textbf{Multi-Agent Gaussian Consensus}}, with intra- and inter-agent Gaussian selection, association, alignment, and optimization, enabling efficient tracking, rapid loop closure and high-fidelity mapping.}}
\label{fig:teaser}\vspace{-2mm}
\end{figure}

\begin{table*}[t!] \centering 
 \centering\setlength{\tabcolsep}{4mm}
\setlength{\abovecaptionskip}{0.1cm} 
\caption{\textbf{Comparison of collaborative multi-agent RGB-D SLAM.} Methods are ranked and highlighted as \colorbox{tabfirst}{\bf first}, \colorbox{tabsecond}{second}, and \colorbox{tabthird}{third}. }\label{tab:sota_method_comp}
\resizebox{\linewidth}{!}{
\begin{tabular}{l|ll|cc}
\toprule
\textbf{Method} & \textbf{3D Map Representation} & \textbf{Map Detail and Completeness} & \textbf{Map Fidelity} & \textbf{Speed} \\
\midrule
CCM-SLAM~\cite{ccmslam} & Sparse keypoint-based maps & Basic level of detail & \colorbox{tabthird}{Low} &  \colorbox{tabfirst}{\bf High} \\
ORB-SLAM3~\cite{orbslam3} & Sparse keypoint-based maps & Basic level of detail & \colorbox{tabthird}{Low} &  \colorbox{tabfirst}{\bf High} \\
Swarm-SLAM~\cite{swarm-slam} & Sparse maps with limited landmarks & Basic level of detail & \colorbox{tabthird}{Low} & \colorbox{tabfirst}{\bf High} \\
CP-SLAM~\cite{hu2024cp} & Dense maps via implicit representation & Moderately complete, yet fragmented in detail & \colorbox{tabsecond}{Mid} & \colorbox{tabthird}{Low} \\
\textbf{MAC-Ego3D} (\textbf{Ours}) & Dense maps with Gaussian splats & Highly detailed and continuous  & \colorbox{tabfirst}{\bf High} & \colorbox{tabfirst}{\bf High} \\
\bottomrule
\end{tabular}}
 \vspace{-4mm}
\end{table*}

Multi-agent collaborative ego-motion estimation and 3D reconstruction, known as multi-agent simultaneous localization and mapping (SLAM), is fundamental for achieving scalable, high-fidelity spatial understanding in applications such as robotic swarms~\cite{8424838} and augmented reality~\cite{holynski2018fast}. In this process~\cite{lajoie2021towards, swarm-slam, orbslam3, tian2022kimera}, each agent independently estimates its pose and constructs a local map, collectively contributing to a shared 3D representation of the environment.

Despite recent efforts to advance multi-agent SLAM, existing systems~\cite{orbslam3, cvislam, ccmslam, swarm-slam, Kimera-Multi} remain limited in scope, primarily focusing on localization and producing sparse, low-fidelity maps that lack the detail needed for photorealistic scene reconstruction. Centralized architectures~\cite{cvislam} struggle with increased latency and communication demands as more agents join, limiting scalability. This has led to more scalable distributed approaches~\cite{swarm-slam, Kimera-Multi}, yet recent multi-agent dense mapping methods~\cite{hu2024cp, matsuki2024gaussian, ha2024rgbd} still face issues of latency and inconsistency. These limitations  highlight  the need for a scalable approach capable of delivering high-fidelity, globally consistent 3D representation in real time.

To address these challenges, we introduce MAC-Ego3D, a collaborative real-time multi-agent SLAM framework that enables online high-fidelity 3D reconstruction through the proposed {\textit{Multi-Agent Gaussian Consensus}} mechanism. As shown in Fig.~\ref{fig:teaser}a, MAC-Ego3D uses Gaussian splats~\cite{gaussiansplatting} to represent the environment with Gaussian primitives that capture spatial structure and appearance attributes, enabling each agent to independently construct and refine local maps with smooth geometry while periodically sharing poses and Gaussian maps for collaborative map and ego-motion optimization.
 MAC-Ego3D has two core components: {\textit{Intra-Agent Gaussian Consensus}}, where each agent builds a local Gaussian-based 3D map from its own observations by leveraging temporal coherence across Gaussian splats, and {\textit{Inter-Agent Gaussian Consensus}}, where multiple agents align overlapping maps by efficiently associating and collaboratively refining co-visible Gaussian splats.

As shown in Fig.~\ref{fig:teaser}b, \textit{Multi-agent Gaussian Consensus} includes four high-level steps: \textbf{1}) {\textit{Selection}}, where agents identify relevant Gaussian splats in overlapping regions to focus alignment efforts, \textbf{2}) {\textit{Association}}, establishing correspondences based on spatial and appearance similarity, \textbf{3}) {\textit{Alignment}}, minimizing spatial discrepancies across overlapping Gaussians, and \textbf{4}) {\textit{Optimization}}, where agents fine-tune their aligned Gaussians and poses to ensure a consistent transformation across local maps.
This consensus mechanism provides MAC-Ego3D with distinct advantages. Leveraging Gaussian splats, each agent models the environment in a continuous representation, enabling smoother, high-resolution 3D reconstructions that surpass the fidelity of traditional feature-based~\cite{orbslam3} or voxel-based SLAM~\cite{hu2024cp}. Additionally, MAC-Ego3D’s parallelized Gaussian consensus design supports independent intra-agent processes and asynchronous inter-agent association, ensuring real-time performance.

Our contributions are summarized as follows:
\begin{itemize}
    \item To the best of our knowledge, we present \textbf{the first real-time full-cycle multi-agent dense SLAM system, \textit{i.e.}, MAC-Ego3D, utilizing Gaussian splat as continuous, photorealistic map representation}, enabling  high-fidelity 3D reconstruction and precise pose tracking.
    \item We propose a \textbf{unified framework for intra-agent and inter-agent Gaussian consensus optimization} that efficiently and collaboratively constrains Gaussian splats, enabling robust loop closure, precise pose estimation, and consistent map alignment.
    \item MAC-Ego3D \textbf{achieves SOTA performance} on several multi-agent benchmarks in both simulated and real-world scenarios, significantly outperforming previous methods in trajectory estimation accuracy (by an order of magnitude on partial sequences), 3D reconstruction fidelity (4–10 dB PSNR improvement), and runtime efficiency ($\times15$ faster than prior SOTA method~\cite{hu2024cp} in pose tracking). 
\end{itemize}

%% file: sec/2_related_work.tex
\section{Related Work}

\noindent\textbf{Collaborative multi-agent RGB-D SLAM} 
\textbf{systems}. These systems~\cite{orbslam3,cvislam,ccmslam} are typically categorized as centralized or distributed architectures. Centralized frameworks, such as CVI-SLAM~\cite{cvislam}, employ a central server to manage data sharing and computational tasks for each agent, handling sub-map management, map fusion, and global bundle adjustment, with processed data relayed back to the agents. CCM-SLAM~\cite{ccmslam} exemplifies this model, where agents perform lightweight visual odometry and transmit localization data and 3D point clouds to the server for refinement.
In contrast, distributed SLAM systems~\cite{swarm-slam,Kimera-Multi} rely on peer-to-peer communication, enabling a decentralized approach with robust outlier rejection. For example, Swarm-SLAM~\cite{swarm-slam} is an open-source, decentralized collaborative SLAM system supporting various sensors with an inter-robot robust loop closure prioritization technique to enhance scalability and flexibility. Building on neural representations in high-fidelity 3D rendering, CP-SLAM~\cite{hu2024cp} leverages NeRF~\cite{nerf}-inspired representation for collaborative mapping, enhancing quality but with mapping times over 10 seconds.
As shown in Table~\ref{tab:sota_method_comp}, MAC-Ego3D advances collaborative SLAM by constraining Gaussian splat~\cite{gaussiansplatting} representation through intra-agent and inter-agent consensus, achieving higher inference speed and 3D reconstruction fidelity.

\begin{figure*}[t!]
\centering   \setlength{\abovecaptionskip}{0.2cm}
\includegraphics[width=\textwidth]{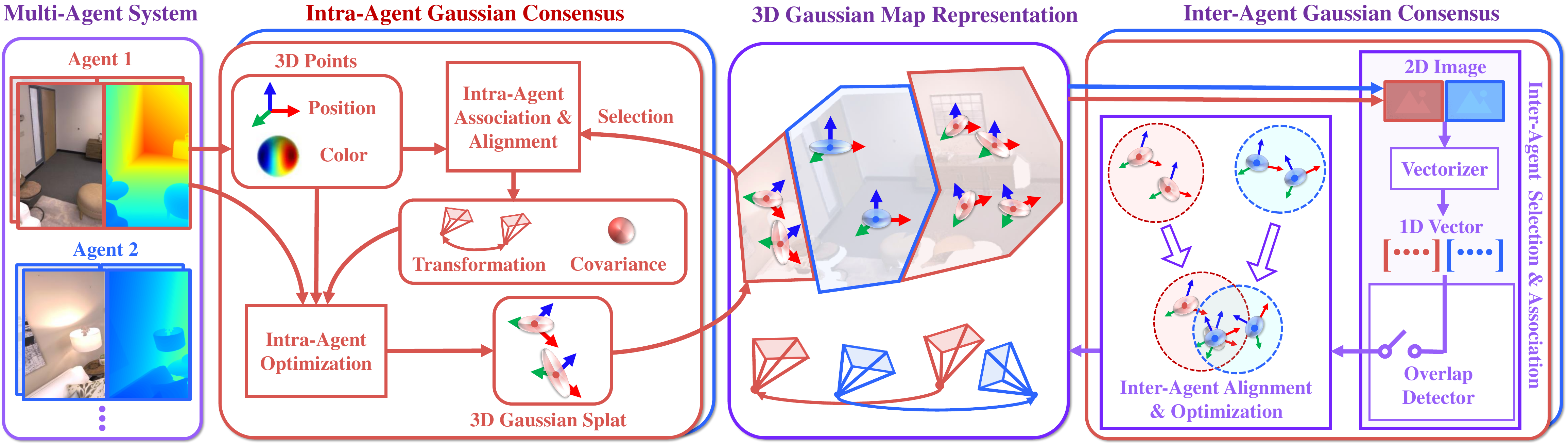}
\caption{\textbf{Pipeline overview of MAC-Ego3D.} MAC-Ego3D leverages parallel {\textit{Intra-Agent Gaussian Consensus}} and periodic {\textit{Inter-Agent Gaussian Consensus}} to enable real-time pose tracking and photorealistic 3D reconstruction using a shared 3D Gaussian map representation.}
\label{fig:pipeline}\vspace{-3mm}
\end{figure*}

\vspace{0.5mm}
\noindent\textbf{\textcolor{black}{Dense RGB-D SLAM models.}} 
Traditional SLAM models like ORB-SLAM~\cite{orbslam,orbslam2} excel in ego-motion estimation but produce sparse maps due to feature-based descriptors. Dense non-neural SLAM models~\cite{kinectfusion,bundlefusion} offer detailed geometry but lack fine appearance detail, while dense neural SLAM models combine geometry and appearance, enabling high-fidelity, photorealistic 3D reconstructions. Implicit neural techniques like NeRF~\cite{rosinol2022nerf} have been incorporated into SLAM for  high-quality textured reconstructions~\cite{imap}, with continued advancements in representation techniques~\cite{coslam,johari2023eslam,niceslam,vox_fusion,point_slam}. Some recent methods also separate learning-based pose estimation from dense mapping~\cite{teed2021droid,zhang2023goslam}. Recently, 3D Gaussian Splatting~\cite{gaussiansplatting} has shown promise for efficient RGB-D rendering, with advancements improving SLAM rendering efficiency and fidelity~\cite{keetha2024splatam, matsuki2024gaussian, ha2024rgbd, ha2025rgbd}. Although some studies~\cite{liso2024loopy, zhu2024loopsplat} have explored enhancing pose optimization through Gaussian splat loop closure, they leave high runtime costs and the challenge of real-time multi-agent mapping largely under-explored, remaining constrained to single-agent systems. MAC-Ego3D bridges this gap by generalizing Gaussian representations to collaborative SLAM through a multi-agent consensus mechanism. It enables agents to independently refine dense local maps while periodically aligning them globally, achieving coherent, high-fidelity photorealistic 3D reconstruction in real time.

%% file: sec/3_method.tex
\section{Method: MAC-Ego3D} \label{sec:method}
\subsection{Problem Formulation}
\noindent\textbf{Collaborative multi-agent  RGB-D SLAM}. 
The goal is to enable a set of agents $\mathcal{A} = \{ a_1, a_2, \dots, a_N \}$ to collaboratively build a shared 3D map. Each agent $a_i \in \mathcal{A}$ independently captures RGB-D observations $\mathbf{Z}_{1:t}^{a_i} = \{ (\mathbf{I}_k^{a_i}, \mathbf{D}_k^{a_i}) \}_{k=1}^t$, where $\mathbf{I}_k^{a_i}$ and $\mathbf{D}_k^{a_i}$ are the RGB image and depth map at time $k$. Based on these observations, each agent estimates its poses $\mathbf{T}_{1:t}^{a_i} \in SE(3)$ and incrementally refines a local map $\mathcal{M}^{a_i}$.
The collaborative ego-motion and 3D reconstruction problem is formulated as a joint optimization of poses and local maps across agents, achieving consensus in overlapping regions to form a consistent global map $\mathcal{M}$. The optimization problem is given by:
\begin{small}
\begin{equation}\label{eq:general_objective} 
\begin{split} 
&\min_{\{\mathbf{T}_{1:t}^{a_i}\}_{i=1}^N, \mathcal{M}}  \sum_{i=1}^N f^{a_i}(\mathbf{T}_{1:t}^{a_i}, \mathcal{M}^{a_i}) \\ 
& + \sum_{(i, j) \in \mathcal{G}} g^{a_i, a_j}(\mathbf{T}_{1:t}^{a_i}, \mathbf{T}_{1:t}^{a_j}, \mathcal{M}^{a_i}, \mathcal{M}^{a_j}), 
\end{split}
\end{equation}
\end{small}
where $f^{a_i}$ represents the local objective function for agent $a_i$, modeling the intra-agent mapping and pose estimation error. The term $g^{a_i, a_j}$ promotes inter-agent alignment between two agents $a_i$ and $a_j$. The set $\mathcal{G} \subseteq \mathcal{A} \times \mathcal{A}$ defines the communication graph, where $(i, j) \in \mathcal{G}$ indicates that agents $a_i$ and $a_j$ are connected in communication.

\subsection{Pipeline Overview}
As shown in Fig.~\ref{fig:pipeline}, MAC-Ego3D leverages a unified 3D Gaussian splat representation (Sec.~\ref{sec:gaussian_splat}), enabling parallel intra-agent processes (localization and mapping) and periodic inter-agent Gaussian associations. This is achieved by modeling the local objective function $f^{a_i}$ in Eq.~\ref{eq:general_objective} through \textit{Intra-Agent Gaussian Consensus} (Sec.~\ref{sec:intraagent_gauss}) and the global objective function $g^{a_i, a_j}$ In Eq.~\ref{eq:general_objective} through \textit{Inter-Agent Gaussian Consensus} (Sec.~\ref{sec:interagent_gauss}). We structure the consensus into four unified steps: \textit{Gaussian Selection},  \textit{Association}, \textit{Alignment}, and \textit{Optimization}. The intra-agent and inter-agent Gaussian consensus threads are implemented via asynchronous multiprocessing (Sec.~\ref{sec:parallel}) to improve efficiency.

\subsection{3D Gaussian Map Representation}
\label{sec:gaussian_splat}

The core of our map representation is a set of oriented Gaussian splats~\cite{gaussiansplatting}, denoted as $\mathbf{G}_i$, which collectively model the 3D environment. Each Gaussian $\mathbf{G}_i$ is defined by its mean position $\mathbf{x}_i$, covariance matrix $\mathbf{\Sigma}_i$, opacity $\lambda_i$, and color $\mathbf{c}_i$. These splats can be rendered onto an image plane from any viewpoint using differentiable rendering techniques. The intensity at a pixel $\mathbf{p}$ is given by:
\begin{equation}\label{eq
} \mathbf{I}(\mathbf{p}) = \sum_{i=1}^{N} \mathbf{c}_i , f_i^{\text{proj}}(\mathbf{p}) \prod_{j=1}^{i-1} \left(1 - f_j^{\text{proj}}(\mathbf{p})\right), \end{equation}
where $f_i^{\text{proj}}(\mathbf{p})$ projects  Gaussian $\mathbf{G}_i$ onto pixel $\mathbf{p}$. 

\subsection{Intra-Agent Gaussian Consensus}
\label{sec:intraagent_gauss}

Each agent $a_i$ aims to minimize its local objective function $f^{a_i}$, which encapsulates the mapping and pose estimation errors based on its own observations $\mathbf{Z}_{1:t}^{a_i}$. This process is modeled through \textit{Intra-Agent Gaussian Consensus}, where the agent integrates its observations into a local map $\mathcal{M}^{a_i}$. 

\vspace{0.5mm}
\noindent\textbf{1) Selection}. At each time step $k$, agent $a_i$ captures RGB-D observations $\mathbf{Z}_k^{a_i} = (\mathbf{I}_k^{a_i}, \mathbf{D}_k^{a_i})$ and projects the depth map $\mathbf{D}_k^{a_i}$ into 3D points $\{\mathbf{x}_{k}^{a_i}\}$. To streamline computations while retaining relevant data, the agent selects candidate Gaussians from its local map $\mathcal{M}^{a_i}$ that are within its current field of view and within a specified distance from its estimated pose $\mathbf{T}_k^{a_i}$. Given the sequential nature of SLAM and the temporal consistency inherent in spatial transformations over time, this neighboring region-based selection focuses on map elements in $\mathcal{M}_{\text{cand}}^{a_i}$ that are more likely to correspond to current observations, effectively reducing computational load and enhancing runtime efficiency.

\vspace{0.5mm}
\noindent\textbf{2) Association.}
The observed 3D points are modeled as Gaussians $\mathbf{G}_{k}^{a_i} \sim \mathcal{N}(\mathbf{x}_{k}^{a_i}, \mathbf{\Sigma}_{k}^{a_i})$.
The agent performs data association between these newly observed Gaussians and the candidate map Gaussians $\mathcal{M}_{\text{cand}}^{a_i}$. Association is based on spatial proximity, resulting in correspondences between observed Gaussians and map Gaussians, \textit{i.e.}, $\{ (\mathbf{G}_{k}^{a_i}, \mathbf{G}_{j}^{\mathcal{M}^{a_i}}) \}$.

\vspace{0.5mm}
\noindent\textbf{3) Alignment.}
For pose tracking, each agent maximizes the likelihood of its observations given its current map. The probabilistic formulation is given by:
\begin{equation}
\begin{split}
& p(\mathbf{T}_{1:t}^{a_i}, \mathcal{M}^{a_i} \mid \mathbf{Z}_{1:t}^{a_i}) = p(\mathcal{M}^{a_i})\, p(\mathbf{T}_0^{a_i}) \\
&\times \prod_{k=1}^t p(\mathbf{T}_k^{a_i} \mid \mathbf{T}_{k-1}^{a_i})\, p(\mathbf{Z}_k^{a_i} \mid \mathbf{T}_k^{a_i}, \mathcal{M}^{a_i}),
\end{split}
\end{equation}
where $p(\mathbf{T}_k^{a_i} \mid \mathbf{T}_{k-1}^{a_i})$  and $p(\mathbf{Z}_k^{a_i} \mid \mathbf{T}_k^{a_i}, \mathcal{M}^{a_i})$ represent the motion model and the observation model, respectively.

\vspace{0.5mm}
The likelihood function for the observed Gaussians is:
\begin{small}
\begin{equation}
p(\mathbf{X}_k^{a_i} \mid \mathcal{M}^{a_i}, \mathbf{T}_k^{a_i}) = \prod_{j} \mathcal{N}\left( \mathbf{T}_k^{a_i} \mathbf{x}_{k}^{a_i};\, \mathbf{x}_j^{\mathcal{M}^{a_i}},\, \mathbf{\Sigma}_j^{\mathcal{M}^{a_i}} \right),
\end{equation}
\end{small}
where $\mathbf{x}_j^{\mathcal{M}^{a_i}}$ and $\mathbf{\Sigma}_j^{\mathcal{M}^{a_i}}$ are the mean and covariance of the corresponding map Gaussians. By minimizing the Mahalanobis distance between transformed observed points and map points, the agent refines its pose $\mathbf{T}_k^{a_i}$, effectively aligning observations to the local map.

\vspace{0.5mm}
\noindent\textbf{4) Optimization}.
The agent updates its local map $\mathcal{M}^{a_i}$ by integrating new Gaussian splats from its observations, adjusting existing Gaussians and adding new ones. Additionally, following prior work~\cite{keetha2024splatam}, the local map is further optimized via differentiable rendering to align with the observed RGB-D images. This involves minimizing a photometric loss between rendered images of the map and the actual observations, refining the Gaussian parameters to enhance map coherence and accuracy. This continuous updating ensures that the local objective $f^{a_i}$ is minimized over time.

\subsection{Inter-Agent Gaussian Consensus}
\label{sec:interagent_gauss}

To construct a consistent global map $\mathcal{M}$, agents perform \textit{Inter-Agent Gaussian Consensus}, aligning overlapping regions of local maps to optimize the global objective $g^{a_i, a_j}$ and ensure agreement in co-visible areas.
 %The inter-agent Gaussian consensus also consists of four steps: Gaussian Selection, Association, Alignment, and Optimization.

\noindent\textbf{1) Selection}.
Unlike intra-agent Gaussian selection, which maintains temporal continuity, inter-agent selection must first establish correspondences across agents. To accomplish this, each agent $a_i$ communicates its embedded observed data—specifically, 2D images $\mathbf{I}^{a_i}$ from previous observations—with other agents to detect potential overlaps at a defined frame interval $T_{comm}$.
%each agent $a_i$ renders its map $\mathcal{M}^{a_i}$ to obtain an image $\mathbf{I}^{a_i}$ and
The image $\mathbf{I}^{a_i}$ is then vectorized into a 1D vector representation $\mathbf{v}^{a_i}$, reducing computational cost and memory usage. Agents compute similarity scores among these compact vectors using a similarity metric $ \mathcal{S}(\mathbf{v}^{a_i}, \mathbf{v}^{a_j}) $ to detect potential overlaps:
\begin{equation}
p(\mathbf{I}^{a_i}, \mathbf{I}^{a_j} \mid \mathcal{H}^{a_i, a_j}) = \mathcal{S}(\mathbf{v}^{a_i}, \mathbf{v}^{a_j}),
\end{equation}
where $\mathcal{H}^{a_i, a_j}$ represents the hypothesized transformation between agents. If $\mathcal{S}(\mathbf{v}^{a_i}, \mathbf{v}^{a_j})$ exceeds a threshold $\tau$, an inter-agent loop closure is detected, prompting agents to select corresponding candidate Gaussian primitives from overlapping regions for further processing.

\vspace{0.5mm}
\noindent\textbf{2) Association.}
Once overlaps are detected, agents associate Gaussians between their maps, establishing inter-agent correspondences $\{ (\mathbf{G}_k^{a_i}, \mathbf{G}_l^{a_j}) \}$.

\vspace{0.5mm}
\noindent\textbf{3) Alignment.}
After association, agents align their maps by finding the transformation $\mathbf{T}^{a_i a_j}$ that minimizes discrepancies between corresponding Gaussians:
\begin{equation}
\begin{split}
\mathbf{T}^{a_i a_j} = & \arg \min_{\mathbf{T}} \sum_{(k, l)} \Big(\left( \mathbf{T} \mathbf{x}_k^{a_i} - \mathbf{x}_l^{a_j} \right)^\top \\
& \times \left( \mathbf{\Sigma}_k^{a_i} + \mathbf{\Sigma}_l^{a_j} \right)^{-1} \left( \mathbf{T} \mathbf{x}_k^{a_i} - \mathbf{x}_l^{a_j} \right)\Big) .
\end{split}
\end{equation}
This process minimizes that the global objective $g^{a_i, a_j}$ by aligning the overlapping regions of the agents' maps.

\vspace{1mm}
\noindent\textbf{4) Optimization.}
After alignment, agents integrate their local maps into a unified global map $\mathcal{M}$. The Gaussian splat representation enables efficient differentiable rendering, allowing gradients to be backpropagated across agents for joint optimization. The optimization penalizes visual and geometric inconsistencies using the loss function:
\begin{small}
\begin{equation}\label{eq:optim} \begin{split} 
 \mathcal{L}(\mathcal{M}) =  &  \sum_{(i,j)}  \Big(\lambda_{\text{L1}} \cdot |\mathbf{I}_i - \mathbf{I}_j |_1 + \lambda_{\text{SSIM}} \cdot \text{SSIM}(\mathbf{I}_i, \mathbf{I}_j) \Big) \\
& + \lambda_{\text{geom}} \sum_{(i,j)} |\mathbf{D}_i - \mathbf{D}_j|_1,
\end{split} \end{equation}
\end{small}
where the first term enforces visual consistency via L1 RGB differences, the second ensures structural consistency with SSIM, and the third promotes geometric alignment by minimizing L1 depth difference, $|\mathbf{D}_i - \mathbf{D}_j|_1$. Depths $\mathbf{D}_i$ and $\mathbf{D}_j$, rendered from Gaussian splats through differentiable rendering, enable gradient-based optimization.

Combining intra-agent and inter-agent Gaussian consensus, we formulate a unified optimization objective integrating map and pose consistency:
\begin{equation}\label{eq:uni_optim}
\mathcal{J} = \mathcal{L}(\mathcal{M}) + \alpha \sum_{(i,j)} \left\| \mathbf{T}^{a_i} \mathbf{T}_{i,j} - \mathbf{T}^{a_j} \right\|_{\Sigma_{i,j}}^2,
\end{equation}
where $\mathbf{T}_{i,j}$ is the relative transformation between agents, and $\alpha$ balances map consistency with pose alignment. By jointly optimizing the poses $\{ \mathbf{T}^{a_i} \}$ and the global map $\mathcal{M}$, agents achieve consistent and accurate environment reconstruction.

\subsection{Gaussian Consensus Parallelization}
\label{sec:parallel}

Our consensus-based framework allows for parallel execution of intra-agent and inter-agent processes.

\noindent\textbf{Intra-Agent.} Each agent independently performs pose estimation and local map updating, as these processes depend solely on local observations and computations. This independence allows for simultaneous processing across all agents.

\noindent\textbf{Inter-Agent.} Inter-agent consensus processes are executed asynchronously to prevent delays in each agent's operation. Differentiable rendering and batch processing enable efficient computation of gradients during global optimization.

%% file: sec/4_experiment.tex
\section{Experiment}
\subsection{Experiment Setup}

\begin{table}[t]
\centering
 \centering \setlength{\tabcolsep}{2.5mm}
\setlength{\abovecaptionskip}{0.1cm}
\caption{\textbf{Multi-agent tracking performance (ATE RMSE$\downarrow$ [cm]) on \textit{Multi-agent Replica} dataset.} A-0, A-1, A-2, and O-0-C refer to the scenes \textit{Apartment-0}, \textit{Apartment-1}, \textit{Apartment-2}, and \textit{Office-0-C}, respectively, where the apartments are multi-room settings and the office is a single-room setting. Best results are highlighted as \colorbox{tabfirst}{\bf first}, \colorbox{tabsecond}{second}, and \colorbox{tabthird}{third}. This result table is extended from~\cite{hu2024cp}.}
\label{tab:multi_agent_tracking}
\resizebox{0.48\textwidth}{!}{
\begin{tabular}{l | c | c c c c | c}
    \toprule
%    \multirowcell{2}{\textbf{Method}} & \multirowcell{2}{\textbf{Setting}} & \multirowcell{2}{\textbf{Apartment-0}\\(\textbf{Multi-room})} & \multirowcell{2}{\textbf{Apartment-1}\\(\textbf{Multi-room})} & \multirowcell{2}{\textbf{Apartment-2}\\(\textbf{Multi-room})} & \multirowcell{2}{\textbf{Office-0-C}\\(\textbf{Single Room})} & \multirowcell{2}{\textbf{Average}} \\  \\  \midrule
       {\textbf{Method}} & {\textbf{Setting}} & {\textbf{A-0}} & {\textbf{A-1}} & {\textbf{A-2}} &{\textbf{O-0-C}}  & {\textbf{Avg.}} \\  
        \midrule
    CCM-SLAM~\cite{ccmslam}   & \multirow{5}{*}{\textbf{Agent 1}}  & FAIL & \cellcolor{tabthird}{2.12} & \cellcolor{tabsecond}{0.51} & 9.84 & - \\
    ORB-SLAM3~\cite{orbslam3}  &        & \cellcolor{tabthird}{0.67} & 4.93 & \cellcolor{tabthird}{1.35} & \cellcolor{tabthird}{0.66} & \cellcolor{tabthird}{1.90} \\
    Swarm-SLAM~\cite{swarm-slam} &        & 1.61 & 4.62 & 2.69 & 1.07 & {2.50} \\
    CP-SLAM~\cite{hu2024cp} &        & \cellcolor{tabsecond}{0.62} & \cellcolor{tabsecond}{1.11} & 1.41 & \cellcolor{tabsecond}{0.50} & \cellcolor{tabsecond}{0.91} \\
    \textbf{MAC-Ego3D} &        & \cellcolor{tabfirst}{\textbf{0.10}} & \cellcolor{tabfirst}{\textbf{0.18}} & \cellcolor{tabfirst}{\textbf{0.10}} & \cellcolor{tabfirst}{\textbf{0.21}} & \cellcolor{tabfirst}{\textbf{0.15}} \\
    \midrule
    CCM-SLAM  & \multirow{5}{*}{\textbf{Agent 2}}  & FAIL & 9.31 & \cellcolor{tabsecond}{0.48} & \cellcolor{tabthird}{0.76} & - \\
    ORB-SLAM3 &         & \cellcolor{tabthird}{1.46} & \cellcolor{tabthird}{4.93} & \cellcolor{tabthird}{1.36} & \cellcolor{tabsecond}{0.54} & \cellcolor{tabthird}{2.07} \\
    Swarm-SLAM &         & 1.98 & 6.50 & 8.53 & 1.76 & 4.69 \\
    CP-SLAM &         & \cellcolor{tabsecond}{1.28} & \cellcolor{tabsecond}{1.72} & 2.41 & 0.79 & \cellcolor{tabsecond}{1.55} \\
    \textbf{MAC-Ego3D}  &         & \cellcolor{tabfirst}{\textbf{0.15}} & \cellcolor{tabfirst}{\textbf{0.20}} & \cellcolor{tabfirst}{\textbf{0.10}} & \cellcolor{tabfirst}{\textbf{0.06}} & \cellcolor{tabfirst}{\textbf{0.13}} \\
    \midrule
    CCM-SLAM & \multirow{5}{*}{\textbf{Average}} & FAIL & 5.71 & \cellcolor{tabsecond}{0.49} & 5.30 & - \\
    ORB-SLAM3 &         & \cellcolor{tabthird}{1.07} & \cellcolor{tabthird}{4.93} & \cellcolor{tabthird}{1.36} & \cellcolor{tabsecond}{0.60} & \cellcolor{tabthird}{1.99} \\
    Swarm-SLAM  &         & 1.80 & 5.56 & 5.61 &{1.42} & 3.60 \\
    CP-SLAM  &         & \cellcolor{tabsecond}{0.95} & \cellcolor{tabsecond}{1.42} & {1.91} & \cellcolor{tabthird}{0.65} & \cellcolor{tabsecond}{1.23} \\
    \textbf{MAC-Ego3D} &         & \cellcolor{tabfirst}{\textbf{0.13}} & \cellcolor{tabfirst}{\textbf{0.19}} & \cellcolor{tabfirst}{\textbf{0.10}} & \cellcolor{tabfirst}{\textbf{0.14}} & \cellcolor{tabfirst}{\textbf{0.14}} \\
    \bottomrule
\end{tabular}}
\vspace{-4mm}
\end{table}

\begin{table}[t]
    \centering
    \setlength{\abovecaptionskip}{0.1cm}
     \caption{\textbf{Multi-agent dense mapping performance, measured by RGB-D rendering quality, on \textit{\textit{Multi-agent Replica}} dataset}.}\label{tab:multi-agent-mapping}
         \setlength{\tabcolsep}{3mm}
    \setlength{\abovecaptionskip}{0.1cm}
\resizebox{0.48\textwidth}{!}{
    \begin{tabular}{l |c | c c c c |c}
        \toprule
       {\textbf{Method}} & {\textbf{Setting}} & {\textbf{A-0}} & {\textbf{A-1}} & {\textbf{A-2}} &{\textbf{O-0-C}}  & {\textbf{Avg.}} \\  
        \midrule
        \multicolumn{7}{c}{\textbf{RGB Rendering Quality (PSNR$\uparrow$ [dB])}} \\ \midrule
        CP-SLAM~\cite{hu2024cp}  &    \multirow{2}{*}{\textbf{Agent 1}}     & 31.93 & 27.87 & 24.34 & 32.91 & 29.26 \\
        \textbf{MAC-Ego3D} &        & \textbf{41.72} &  \textbf{36.54} & \textbf{37.91} & \textbf{43.09} & \textbf{39.82} \\
        \midrule
        CP-SLAM &     \multirow{2}{*}{\textbf{Agent 2}}     & 32.39 & 26.97 & 25.90 & 32.33 & 29.40 \\
        \textbf{MAC-Ego3D} &        & \textbf{43.94}&   \textbf{36.02} & \textbf{39.26} & \textbf{41.76} & \textbf{40.25} \\
        \midrule
        CP-SLAM  &  \multirow{2}{*}{\textbf{Average}}   & 32.16 & 27.42 & 25.12 & 32.62 & 29.33 \\
        \textbf{MAC-Ego3D} &        & \textbf{42.83} & \textbf{36.28} & \textbf{38.59} & \textbf{42.43} & \textbf{40.04} \\
       \midrule
        \multicolumn{7}{c}{\textbf{RGB Rendering Quality (SSIM$\uparrow$)}} \\ \midrule
        CP-SLAM~\cite{hu2024cp}  &    \multirow{2}{*}{\textbf{Agent 1}}     & 0.907 & 0.784 & 0.770 & 0.905 & 0.842 \\
        \textbf{MAC-Ego3D} &        & \textbf{0.976} &  \textbf{0.970} & \textbf{0.973} & \textbf{0.985} & \textbf{0.976} \\
        \midrule
        CP-SLAM &     \multirow{2}{*}{\textbf{Agent 2}}     & 0.917 & 0.781 & 0.772 & 0.901 & 0.843 \\
        \textbf{MAC-Ego3D} &        & \textbf{0.985}&   \textbf{0.967} & \textbf{0.978} & \textbf{0.981} & \textbf{0.978} \\
        \midrule
        CP-SLAM  &  \multirow{2}{*}{\textbf{Average}}  & 0.912 & 0.783 & 0.771 & 0.903 & 0.843 \\
        \textbf{MAC-Ego3D} &        & \textbf{0.981} & \textbf{0.969} & \textbf{0.976} & \textbf{0.983} & \textbf{0.977} \\
        \midrule
        \multicolumn{7}{c}{\textbf{RGB Rendering Quality (LPIPS$\downarrow$)}} \\ \midrule
        CP-SLAM~\cite{hu2024cp}  &    \multirow{2}{*}{\textbf{Agent 1}}     & 0.240 & 0.351 & 0.376 & 0.292 & 0.315 \\
        \textbf{MAC-Ego3D} &        & \textbf{0.049} &  \textbf{0.061} & \textbf{0.059} & \textbf{0.030} & \textbf{0.050} \\
        \midrule
        CP-SLAM &     \multirow{2}{*}{\textbf{Agent 2}}     & 0.235 & 0.366 & 0.368 & 0.280 & 0.312 \\
        \textbf{MAC-Ego3D} &        & \textbf{0.049}&   \textbf{0.063} & \textbf{0.051} & \textbf{0.035} & \textbf{0.050} \\
        \midrule
        CP-SLAM  &  \multirow{2}{*}{\textbf{Average}}& 0.238 & 0.359 & 0.372 & 0.286 & 0.314 \\
        \textbf{MAC-Ego3D} &        & \textbf{0.049} & \textbf{0.062} & \textbf{0.055} & \textbf{0.033} & \textbf{0.050} \\
        \midrule
        \multicolumn{7}{c}{\textbf{Depth Rendering Quality (L1 Loss$\downarrow$ [mm])}} \\ \midrule
        CP-SLAM~\cite{hu2024cp}  &    \multirow{2}{*}{\textbf{Agent 1}}     & 0.49 & 1.60 & 3.58 & 0.42 & 1.52 \\
        \textbf{MAC-Ego3D} &        & \textbf{0.48} &  \textbf{1.06} & \textbf{0.92} & \textbf{0.37} & \textbf{0.71} \\
        \midrule
        CP-SLAM &     \multirow{2}{*}{\textbf{Agent 2}}     & 0.76 & 1.31 & 2.56 & 0.46 & 1.27 \\
        \textbf{MAC-Ego3D} &        & \textbf{0.48}&   \textbf{1.00} & \textbf{1.14} & \textbf{0.44} & \textbf{0.77} \\
        \midrule
        CP-SLAM  &  \multirow{2}{*}{\textbf{Average}} & 0.63 & 1.46 & 3.07 & 0.44 & 1.40 \\
        \textbf{MAC-Ego3D} &        & \textbf{0.48} & \textbf{1.03} & \textbf{1.03} & \textbf{0.41} & \textbf{0.74} \\
        \bottomrule
    \end{tabular}}
   \vspace{-5mm}
\end{table}

\begin{figure*}[t!]
\centering   \setlength{\abovecaptionskip}{0.1cm}
\includegraphics[width=\textwidth]{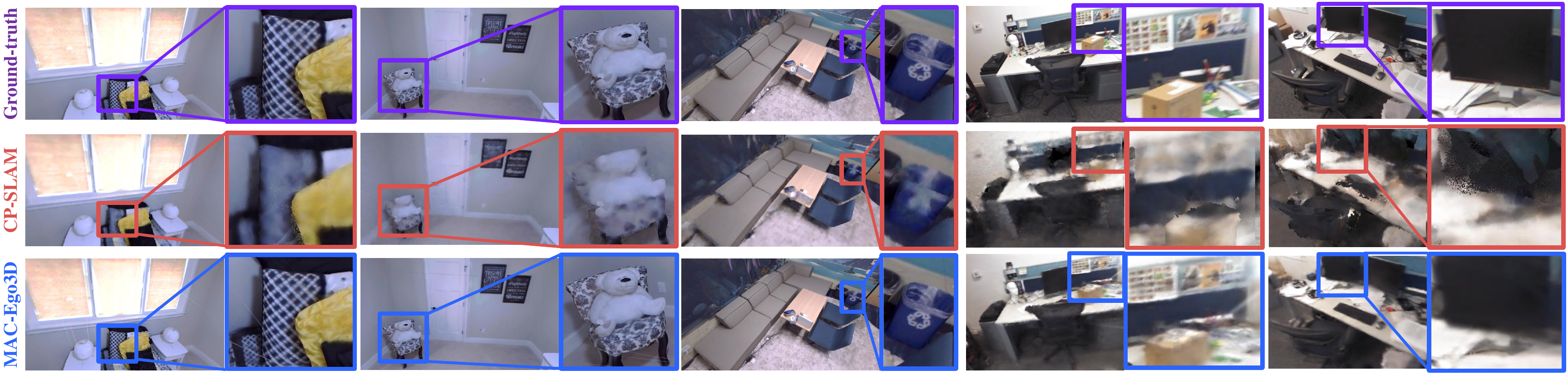}
\caption{\textbf{Qualitative RGB image rendering quality comparison between multi-agent SLAM models with dense reconstruction capability}, \textit{i.e.}, CP-SLAM and our MAC-Ego3D, on \textit{Multi-agent Replica} (\textbf{Left}) and \textit{7-Scenes} (\textbf{Right}) datasets.}
\label{fig:rendering-comparison}\vspace{-4mm}
\end{figure*}
\begin{table}[t]
    \centering
    \setlength{\tabcolsep}{1.8mm}
    \setlength{\abovecaptionskip}{0.1cm}
      \caption{\textbf{Multi-agent tracking and mapping performance on  the real-world \textit{7-Scenes} dataset.} `\textcolor{red}{FAIL}'  indicates partial pose tracking failure. `RedK' denotes the `RedKitchen' sequence.}\label{tab:7scenes-results}
\resizebox{0.48\textwidth}{!}{
    \begin{tabular}{l | c | c c c c c c |c}
        \toprule
       {\textbf{Method}} & {\textbf{Setting}} & {\textbf{Chess}} & {\textbf{Fire}} & {\textbf{Heads}} & {\textbf{Office}} &  {\textbf{{RedK}}} & {\textbf{Stairs}} & {\textbf{Avg.}} \\  
        \midrule
        \multicolumn{9}{c}{\textbf{Tracking Performance (ATE RMSE$\downarrow$ [cm])}} \\ \midrule
        CP-SLAM~\cite{hu2024cp}  &    \multirow{2}{*}{\textbf{Agent 1}}     & \textcolor{red}{FAIL} & 1.90 & 22.11 & \textbf{3.49} & 7.01 & 8.88  & - \\
        \textbf{MAC-Ego3D} &        & \textbf{6.36} &  \textbf{1.63} & \textbf{1.42}  & {4.45} &  \textbf{2.74} & \textbf{2.94} & \textbf{3.26}  \\
        \midrule
        CP-SLAM &     \multirow{2}{*}{\textbf{Agent 2}}     & \textcolor{red}{FAIL} & 2.02 & 9.89 & 8.52 &  \textcolor{red}{FAIL} & \textcolor{red}{FAIL} & - \\ 
        \textbf{MAC-Ego3D} &        & \textbf{10.71} &  \textbf{1.00} & \textbf{9.71} & \textbf{5.85} & \textbf{17.71} & \textbf{3.80} & \textbf{8.13}  \\
        \midrule
        CP-SLAM  &  \multirow{2}{*}{\textbf{Average}}        & \textcolor{red}{FAIL} & 1.96 & 16.00 & 6.01 & \textcolor{red}{FAIL} & \textcolor{red}{FAIL} & - \\
        \textbf{MAC-Ego3D} &        & \textbf{8.54} & \textbf{1.32} & \textbf{5.57} & \textbf{5.20}  & \textbf{10.23} & \textbf{3.37} & \textbf{5.70}  \\
        \midrule
        \multicolumn{9}{c}{{\textbf{RGB Rendering Quality (PSNR$\uparrow$ [dB])}}} \\ \midrule
        CP-SLAM~\cite{hu2024cp}  &    \multirow{2}{*}{\textbf{Agent 1}}     & 15.18 & 18.60 & 17.97 & 19.72 &  18.75 & 16.69 & 17.82 \\
        \textbf{MAC-Ego3D} &        & \textbf{19.42} &  \textbf{20.75} & \textbf{20.66} & \textbf{22.21} &  \textbf{21.25} & \textbf{23.30} & \textbf{21.27} \\
        \midrule
        CP-SLAM &     \multirow{2}{*}{\textbf{Agent 2}}     & 15.87 & 17.38 & 19.76 & 19.68 &  10.00 & 16.81 & 16.58 \\
        \textbf{MAC-Ego3D} &        & \textbf{19.27} &  \textbf{20.47} & \textbf{20.42} & \textbf{20.62} &  \textbf{20.53} & \textbf{23.32} & \textbf{20.77} \\
        \midrule
        CP-SLAM  &  \multirow{2}{*}{\textbf{Average}}        & 15.53 & 17.99 & 18.87 & 19.70 &  14.38 & 16.75 & 17.20 \\
        \textbf{MAC-Ego3D} &        & \textbf{19.35} & \textbf{20.61} & \textbf{20.54} & \textbf{21.42} &  \textbf{20.89} & \textbf{23.31} & \textbf{21.02} \\
        \midrule
        \multicolumn{9}{c}{{\textbf{RGB Rendering Quality (SSIM$\uparrow$)}}} \\ \midrule
        CP-SLAM~\cite{hu2024cp}  &    \multirow{2}{*}{\textbf{Agent 1}}     & 0.516 & 0.568 & 0.652 & 0.695 &  0.621 & 0.580 & 0.605 \\
        \textbf{MAC-Ego3D} &        & \textbf{0.738} &  \textbf{0.694} & \textbf{0.811} & \textbf{0.832} & \textbf{0.765} & \textbf{0.814} & \textbf{0.776} \\
        \midrule
        CP-SLAM &     \multirow{2}{*}{\textbf{Agent 2}}     & 0.538 & 0.541 & 0.706 & 0.705 &  0.374 & 0.612 & 0.579 \\
        \textbf{MAC-Ego3D} &        & \textbf{0.752} & \textbf{0.701} & \textbf{0.794} & \textbf{0.819} & \textbf{0.725} & \textbf{0.836} & \textbf{0.771} \\
        \midrule
        CP-SLAM  &  \multirow{2}{*}{\textbf{Average}}        & 0.527 & 0.555 & 0.555 & 0.679 &  0.498 & 0.596 & 0.592 \\
        \textbf{MAC-Ego3D} &        & \textbf{0.745} & \textbf{0.698} & \textbf{0.803} & \textbf{0.826} &  \textbf{0.745} & \textbf{0.825} & \textbf{0.774} \\
        \midrule
        \multicolumn{9}{c}{\textbf{RGB Rendering Quality (LPIPS$\downarrow$)}} \\ \midrule
        CP-SLAM~\cite{hu2024cp} &     \multirow{2}{*}{\textbf{Agent 1}}     & 0.558 & 0.546 & 0.532 & 0.502 &  0.536 & 0.542 & 0.536 \\
        \textbf{MAC-Ego3D} &        & \textbf{0.322} &  \textbf{0.309} & \textbf{0.258} & \textbf{0.236} & \textbf{0.319} & \textbf{0.207} & \textbf{0.275} \\
        \midrule
        CP-SLAM &     \multirow{2}{*}{\textbf{Agent 2}}     & 0.557 & 0.545 & 0.489 & 0.493 & 0.710 & 0.542 & 0.556 \\
        \textbf{MAC-Ego3D} &        & \textbf{0.300} &  \textbf{0.309} & \textbf{0.285} & \textbf{0.268} &  \textbf{0.371} & \textbf{0.210} & \textbf{0.291} \\
        \midrule
        CP-SLAM  &  \multirow{2}{*}{\textbf{Average}}        & 0.558 & 0.546 & 0.511 & 0.498 &  0.623 & 0.542 & 0.546 \\
        \textbf{MAC-Ego3D} &        & \textbf{0.311} & \textbf{0.309} & \textbf{0.272} & \textbf{0.252} &  \textbf{0.345} & \textbf{0.209} & \textbf{0.283} \\
        \bottomrule
    \end{tabular}}
  \vspace{-3.5mm}
\end{table}

\noindent\textbf{Datasets}. We use two multi-agent datasets for evaluation:
\begin{itemize}
    \item \textbf{Synthetic \textit{Multi-agent Replica} dataset}~\cite{replica, hu2024cp} offers high-quality RGB-D sequences with eight collections—four single-agent trajectories (1,950 frames each) and four multi-agent SLAM scenarios. Collaborative sequences are divided by agent, with 2,500 frames each, except for the sequence \textit{Office-0-C}, which has 1,500 frames.
    \item \textbf{Real-world \textit{7-Scenes} dataset}~\cite{7scene} includes real-world indoor scenes with multiple handheld RGB-D sequences of 500 or 1,000 frames each. It challenges tracking and reconstruction due to the presence of degraded visual observation, invalid depth values, and complex motion.
\end{itemize}

\noindent\textbf{Evaluation metrics.} For trajectory tracking, we use Absolute Trajectory Error (ATE) measured by Root Mean Squared Error (RMSE), denoted as ATE RMSE ($\downarrow$). Mapping quality is evaluated with RGB rendering metrics—PSNR ($\uparrow$), SSIM ($\uparrow$), and LPIPS ($\downarrow$)—and depth rendering accuracy through depth L1 loss ($\downarrow$), where $\downarrow$ and $\uparrow$ indicate that lower and higher values, respectively, signify better performance.
%\textcolor{red}{3D mesh recon?}}

\noindent\textbf{Baseline methods}.
To evaluate our method in multi-agent scenarios, we benchmark against diverse collaborative SLAM approaches, including classical methods (CCM-SLAM~\cite{ccmslam}, Swarm-SLAM~\cite{swarm-slam}, ORB-SLAM3~\cite{orbslam3}) and the neural-based CP-SLAM~\cite{hu2024cp}. This range of traditional and neural SLAM systems ensures robust comparison. In experiments on the \textit{Multi-agent Replica} dataset with loop-closure data, we follow~\cite{hu2024cp} to compare with dense SLAM systems (NICE-SLAM~\cite{niceslam}, Vox-Fusion~\cite{vox-fusion}) and ORB-SLAM3~\cite{orbslam3}.

\noindent\textbf{Implementation details.}
For inter-agent candidate selection, we use an off-the-shelf pre-trained feature embedding model~\cite{Izquierdo_CVPR_2024_SALAD} with dot product as the similarity metric $\mathcal{S}$ and a threshold $\tau$ of 0.8. Point selection in Generalized-ICP~\cite{segal2009generalized} follows a distance threshold of 0.02. In the global Gaussian map optimization loss function (Eq.~\ref{eq:optim}), the weights $\lambda_{\text{L1}}$, $\lambda_{\text{SSIM}}$, and $\lambda_{\text{geom}}$ are set to 0.8, 0.2, and 0.1, respectively, with the coefficient $\alpha$ in Eq.~\ref{eq:uni_optim} set to 1. The inter-agent communication interval $T_{comm}$ is set as 150 and  60 for \textit{Multi-agent} \textit{Replica} and \textit{7-Scenes}, respectively. Experiments were conducted on a desktop with an Intel Xeon Gold 5318Y CPU and NVIDIA RTX 6000 Ada GPUs running Ubuntu 20.04. Our method was implemented in Python using the PyTorch framework. Additional details are provided in the appendix.

\begin{figure}[t!]
\centering
\includegraphics[width=0.48\textwidth]{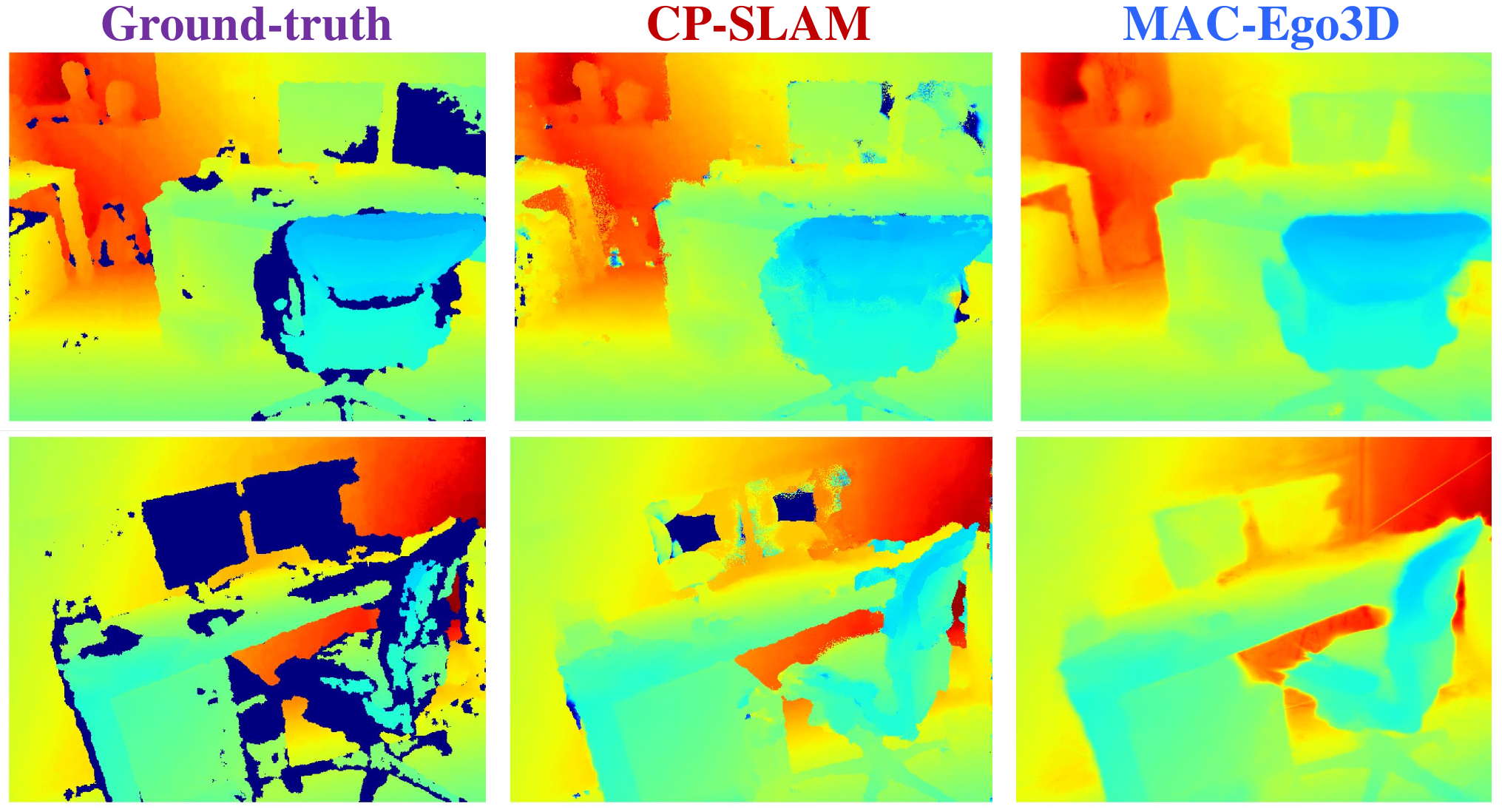}\vspace{-2mm}
\caption{\textbf{Qualitative depth rendering comparison on  \textit{7-Scenes}. } } 
%\vspace{-1mm}
\label{fig:depth_rendering}\vspace{-4mm}
\end{figure}

\subsection{Main Quantitative Results} 

\textcolor{black}{
\noindent\textbf{Tracking results on \textit{Multi-agent Replica} dataset}. 
As shown in Table~\ref{tab:multi_agent_tracking}, MAC-Ego3D consistently achieves the lowest tracking errors across all cases and agents, significantly outperforming classical multi-agent sparse mapping methods and the dense CP-SLAM. It improves average trajectory estimation accuracy by nearly an order of magnitude over the previous SOTA method (0.14 cm vs. 1.23 cm in ATE). This exceptional performance is driven by the effective establishment of visual and geometric consensus across the Gaussian splat map representation of multiple agents. }

%\vspace{1mm}
\noindent\textbf{Dense mapping results on \textit{Multi-agent Replica} dataset.} Table~\ref{tab:multi-agent-mapping} highlights MAC-Ego3D’s exceptional dense mapping performance, achieving 40.04 dB in PSNR—over 10 dB higher than CP-SLAM’s 29.33 dB. It also delivers significant gains in SSIM (0.977 vs. 0.843) and LPIPS (0.050 vs. 0.314), reflecting superior visual fidelity and texture preservation. Furthermore, MAC-Ego3D halves the depth rendering error, reducing it to 0.74 mm compared to CP-SLAM's 1.40 mm. These results verify MAC-Ego3D’s superior accuracy in modeling geometry and appearance in 3D reconstruction.

%\vspace{1mm}
\noindent\textcolor{black}{
\textbf{Tracking and dense mapping results on real-world \textit{7-Scenes} dataset}.
In Table~\ref{tab:7scenes-results}, we compare two multi-agent SLAM models with dense reconstruction capabilities, CP-SLAM and our MAC-Ego3D, on the challenging \textit{7-Scenes} dataset. MAC-Ego3D demonstrates clear superiority, particularly in complex scenes, achieving significantly lower trajectory estimation error (ATE) across most sequences. It accurately track poses in sequences like `Chess' and `Stairs', where CP-SLAM often encounters failures or drift. For RGB rendering quality, MAC-Ego3D consistently outperforms CP-SLAM, achieving a 4 dB improvement in PSNR, a 31\% increase in SSIM, and nearly halving the LPIPS score.}

\begin{figure}[t!]
\centering
\includegraphics[width=0.48\textwidth]{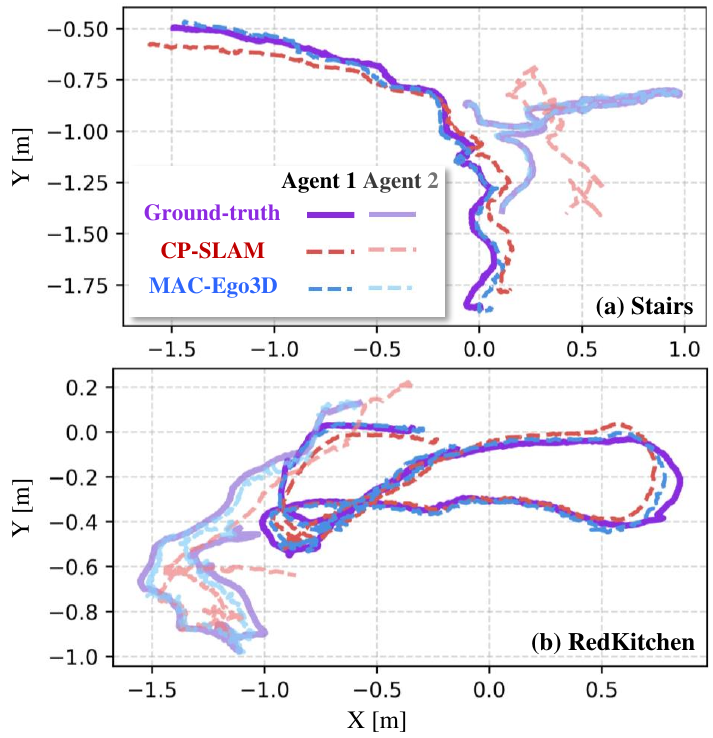}\vspace{-3mm}
\caption{\textbf{Multi-agent trajectory estimation results on  \textit{7-Scenes}. } } 
\vspace{-1mm}
\label{fig:trajectory_estimation}\vspace{-5mm}
\end{figure}

\begin{table}[t]
\centering \centering\setlength{\tabcolsep}{1mm}
\setlength{\abovecaptionskip}{0.1cm} 
\caption{\textbf{Performance and runtime of sequences with loop  on \textit{\textit{Multi-agent Replica}}.} We compare  ATE RMSE$\downarrow$ [cm] alongside per-frame tracking and mapping times, extending results from~\cite{hu2024cp}.}
\resizebox{\linewidth}{!}{
\begin{tabular}{l || c c c c | c || c | c}
    \toprule
    \textbf{Method} & \textbf{R0} & \textbf{R1} & \textbf{O0} & \textbf{O3} & \textbf{Avg.} & \textbf{Track} & \textbf{Map} \\
    \midrule
  
    NICE-SLAM~\cite{niceslam} & 1.27 & 1.74 & 2.27 & 3.19 & 2.12 & 1.77s & 11.26s \\
    Vox-Fusion~\cite{vox-fusion} & 0.82 & 1.35 & 0.99 & \cellcolor{tabthird}0.82 & 0.99 & 0.36s & \cellcolor{tabthird}0.83s \\ \midrule
      ORB-SLAM3~\cite{orbslam3} & \cellcolor{tabthird}0.54 & \cellcolor{tabsecond}0.21 & \cellcolor{tabthird}0.58 & 0.89 & \cellcolor{tabthird}0.56 & \cellcolor{tabsecond}0.03s & \cellcolor{tabsecond}0.32s \\
    CP-SLAM~\cite{hu2024cp} & \cellcolor{tabsecond}0.48 & \cellcolor{tabthird}0.44 & \cellcolor{tabsecond}0.56 & \cellcolor{tabsecond}0.37 & \cellcolor{tabsecond}0.46 & \cellcolor{tabthird}0.30s & 10.10s \\
    \textbf{MAC-Ego3D} & \cellcolor{tabfirst}\textbf{0.13} & \cellcolor{tabfirst}\textbf{0.17} & \cellcolor{tabfirst}\textbf{0.16} & \cellcolor{tabfirst}\textbf{0.14} & \cellcolor{tabfirst}\textbf{0.15} & \cellcolor{tabfirst}\textbf{0.02s} & \cellcolor{tabfirst}{\textbf{0.05s}} \\
    \bottomrule
\end{tabular}}
\vspace{-5mm}
\label{tab:performance_runtime_analysis}
\end{table}

\subsection{Qualitative Results}

\noindent\textcolor{black}{
\textbf{RGB rendering}.
Fig.~\ref{fig:rendering-comparison} showcases MAC-Ego3D’s superior RGB rendering quality compared to CP-SLAM in multi-agent dense reconstruction. MAC-Ego3D closely aligns with ground truth, capturing fine details and accurate colors, while CP-SLAM exhibits noticeable artifacts, particularly in high-frequency textured areas, \textit{e.g.}, the stripes of the cushion.}

\vspace{0.5mm}
\noindent\textbf{Depth rendering}. Fig.~\ref{fig:depth_rendering} shows that the reconstructed 3D map of MAC-Ego3D renders high-quality, continuous depth maps that accurately capture 3D scene geometry. In contrast to CP-SLAM, which often produces incomplete or fragmented geometry, MAC-Ego3D ensures depth completeness even in challenging regions where sensor limitations and reconstruction errors typically cause failures.

\vspace{0.5mm}
\noindent\textcolor{black}{
\textbf{Trajectory estimation}. Fig.~\ref{fig:trajectory_estimation} illustrates the superior alignment of MAC-Ego3D with ground-truth trajectories, particularly during sharp turns, consistently maintaining accurate poses across agents. In contrast, CP-SLAM shows significant drift, with MAC-Ego3D capturing intricate path details more precisely in all evaluated scenes.
}

\subsection{Efficiency and Robustness Analyses}
%\noindent\textcolor{black}{\textbf{Runtime efficiency and performance on repetitive loops.}
\noindent\textbf{Runtime evaluation.} \textcolor{black}{Table~\ref{tab:performance_runtime_analysis} shows the runtime efficiency and superior accuracy of MAC-Ego3D in sequences with repetitive loops of \textit{Multi-agent Replica}. Compared to CP-SLAM, which employs implicit neural representations for dense reconstruction but demands high computational time, MAC-Ego3D achieves lower errors with an average ATE of {0.15 cm} and outperforms baseline models in runtime, with tracking and mapping time of {0.02}s and {0.05}s per frame. This efficiency is partly due to the Gaussian splat-based representation, which allows rapid rendering (0.005 s per frame) that accelerates the transformation from Gaussian splats to RGB-D images for inter-agent consensus optimization.}

\begin{figure}[t!]
\centering
\includegraphics[width=0.48\textwidth]{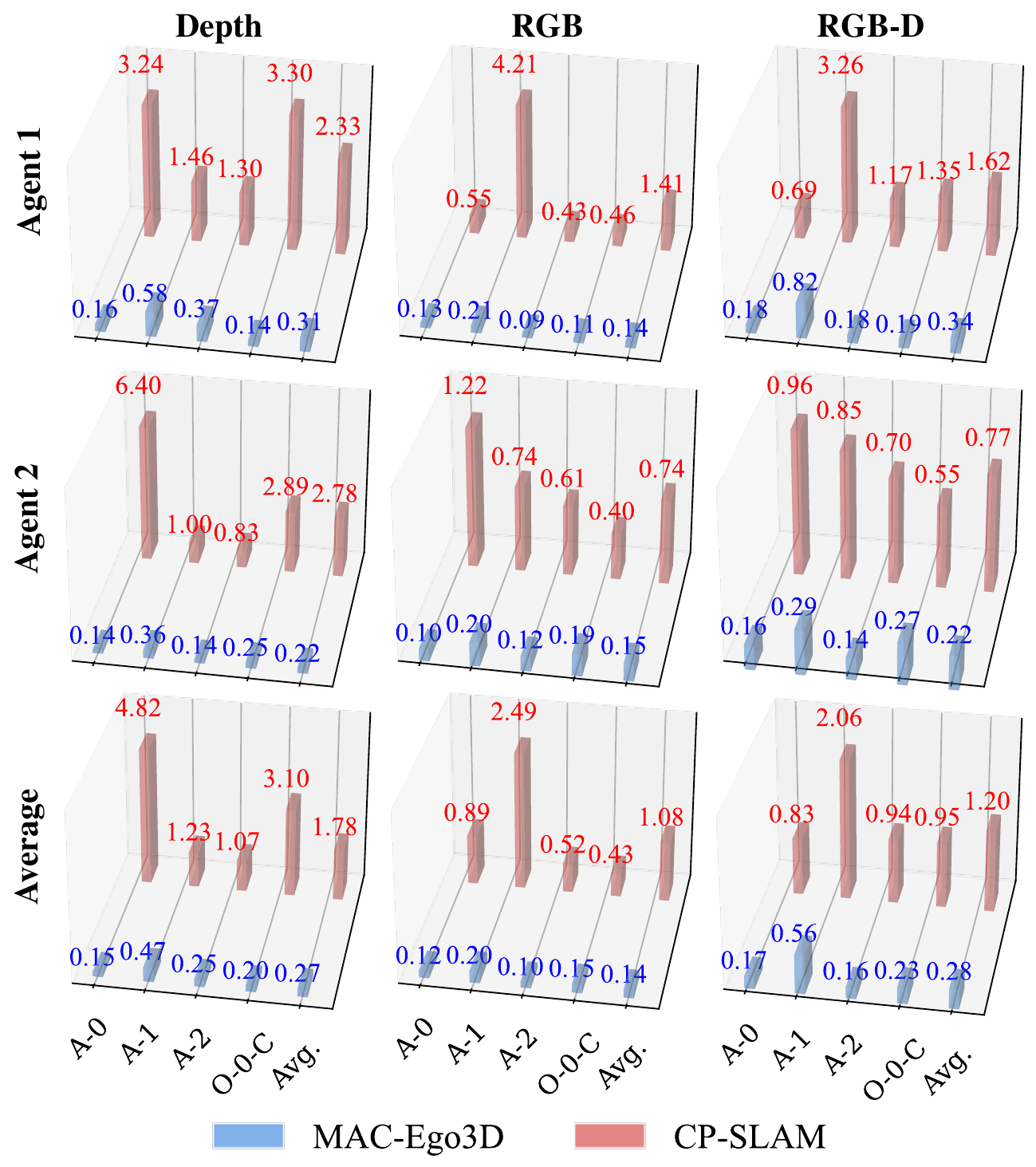}\vspace{-2mm}
\caption{\textbf{Multi-agent tracking robustness (ATE RMSE$\downarrow$ [cm]) under corruptions on perturbed \textit{\textit{Multi-agent Replica}} dataset.} } 
\vspace{-2mm}
\label{tab:robustness}\vspace{-2mm}
\end{figure}

\vspace{0.5mm}
\noindent{\textbf{Robustness under observation corruptions}}.
 \textcolor{black}{As shown in Fig.~\ref{tab:robustness}, MAC-Ego3D consistently outperforms CP-SLAM across challenging visual corruptions, achieving remarkable resilience. Following~\cite{xu2024customizable}, a Gaussian noise model was applied with an RGB color standard deviation of 10 in 255 and a depth standard deviation of 5 mm.  MAC-Ego3D achieves an ATE of 0.27 cm under depth noise, while CP-SLAM's errors are much higher (1.78 cm in ATE). With RGB corruption, MAC-Ego3D maintains a low ATE of 0.14 cm, well below CP-SLAM’s 1.08 cm. Even under composite RGB-D noise, MAC-Ego3D holds steady with an ATE of 0.28 cm, highlighting its robustness under noisy observations.}

\begin{figure}[t!]
\centering \setlength{\abovecaptionskip}{0.2cm}
\includegraphics[width=0.48\textwidth]{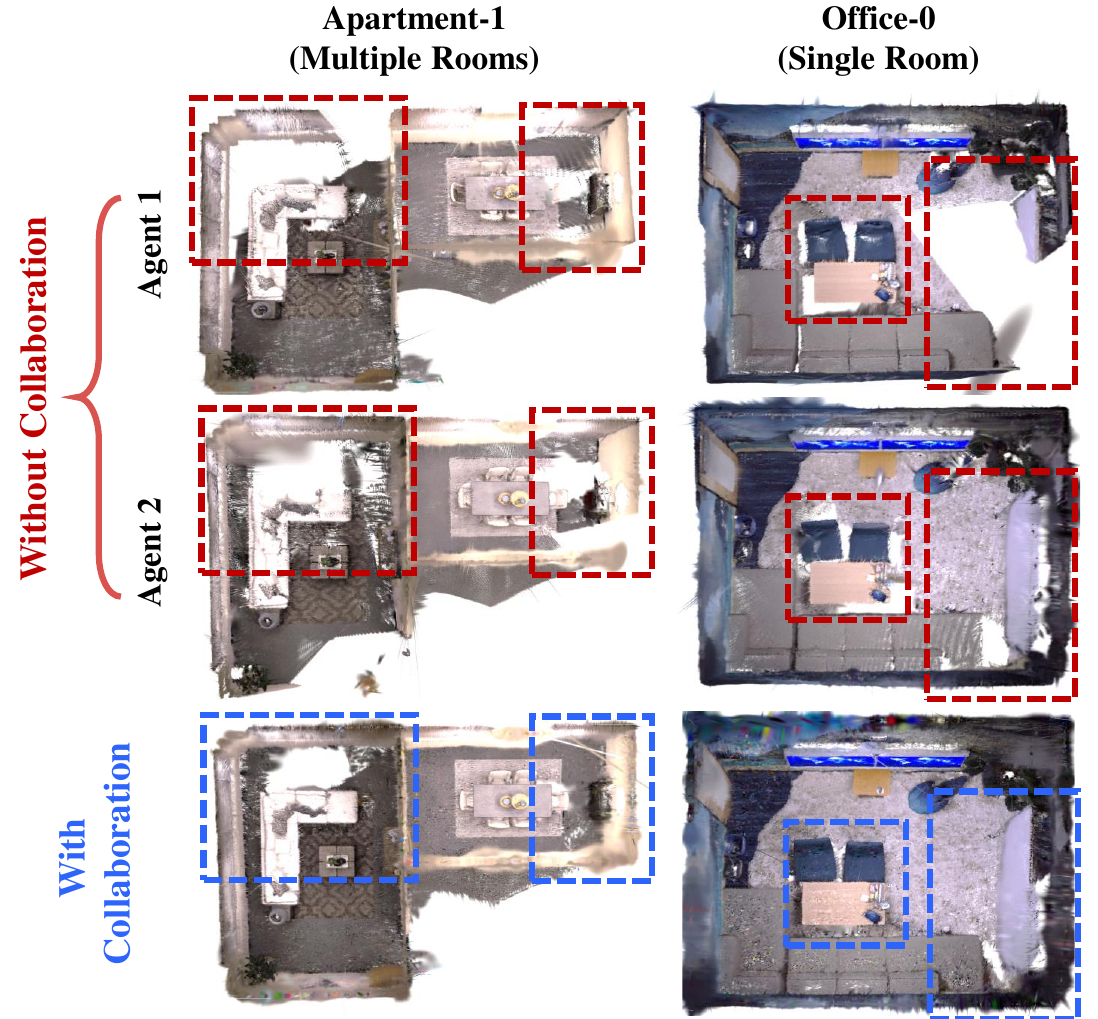}
\caption{\textbf{Real-time 3D reconstruction by our MAC-Ego3D using Gaussian splat representation on \textit{Multi-agent Replica}. } } 
\label{fig:mesh-comparison}\vspace{-2mm}
\end{figure}

\subsection{Ablation Study}

\textcolor{black}{
\noindent\textbf{High-fidelity 3D reconstruction through multi-agent collaboration.} Fig.~\ref{fig:mesh-comparison} illustrates the effectiveness of collaborative mapping in MAC-Ego3D, showing its advantage in producing high-fidelity 3D maps. While individual agents struggle with map consistency and quality, collaboration allows seamless sharing of spatial information, leading to cohesive and accurately aligned reconstructions.
}

\begin{figure}[t!]
\centering 
\begin{subfigure}{0.48\textwidth}
    \centering
    \includegraphics[width=\textwidth]{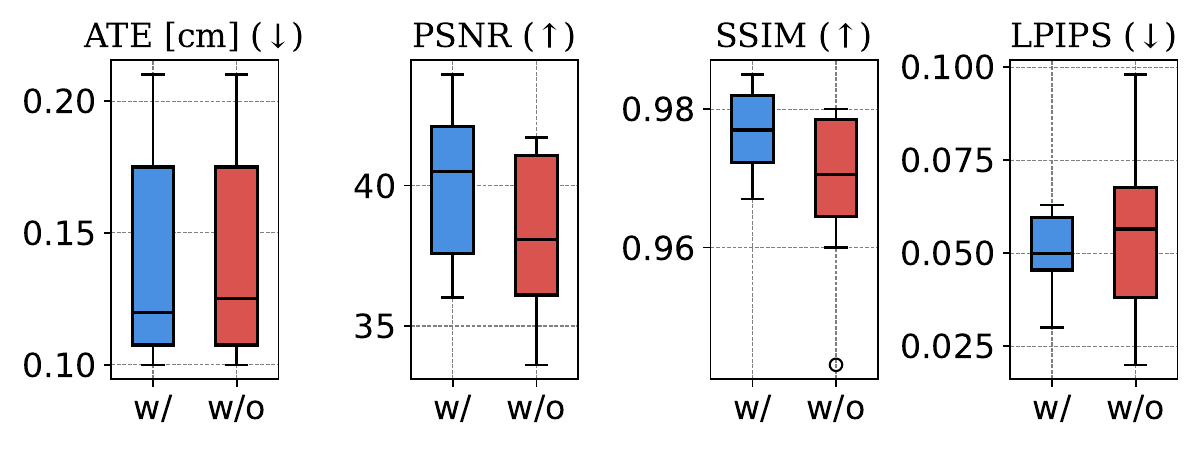}\vspace{-3mm}
    \phantomcaption  % Use \phantomcaption to avoid adding automatic (a) or (b) parentheses
    \label{fig:ma_sa_ablation_replica}
    \textbf{(a)} \textit{Multi-agent Replica}
\end{subfigure}
\hfill
% Subfigure 2: \textit{7Scenes} dataset
\begin{subfigure}{0.48\textwidth}
    \centering
    \includegraphics[width=\textwidth]{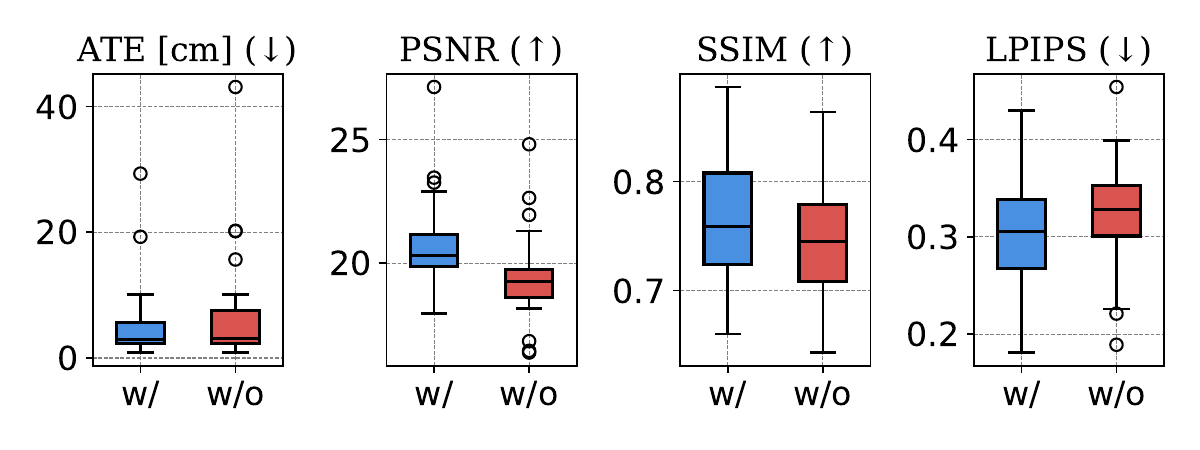}\vspace{-3mm}
    \phantomcaption  % Use \phantomcaption to avoid adding automatic (a) or (b) parentheses
    \label{fig:ma_sa_ablation_7scenes}
    \textbf{(b)} \textit{7-Scenes}
\end{subfigure}\vspace{-3mm}
\caption{\textbf{Ablation study on collaborative 3D map optimization.}}
\label{fig:ma_sa_ablation_combined} \vspace{0.5mm}
\centering %\setlength{\abovecaptionskip}{0.5cm}
% Subfigure 1: \textit{Multi-agent Replica} dataset
\begin{subfigure}{0.48\textwidth}
    \centering
    \includegraphics[width=\textwidth]{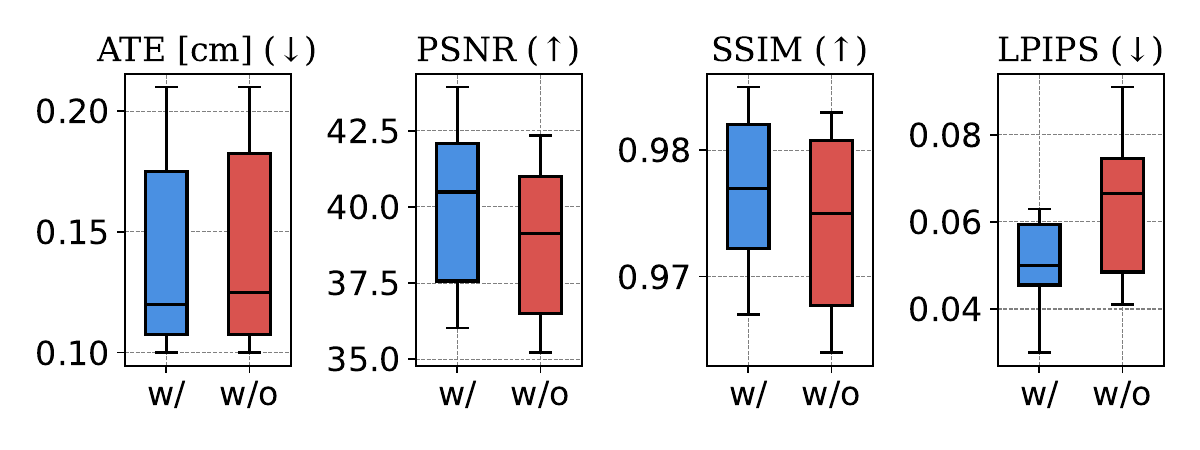}\vspace{-3mm}
    \phantomcaption  % Use \phantomcaption to avoid adding automatic (a) or (b) parentheses
    \textbf{(a)} \textit{Multi-agent Replica}
\end{subfigure}
\hfill
% Subfigure 2: \textit{7Scenes} dataset
\begin{subfigure}{0.48\textwidth}
    \centering
    \includegraphics[width=\textwidth]{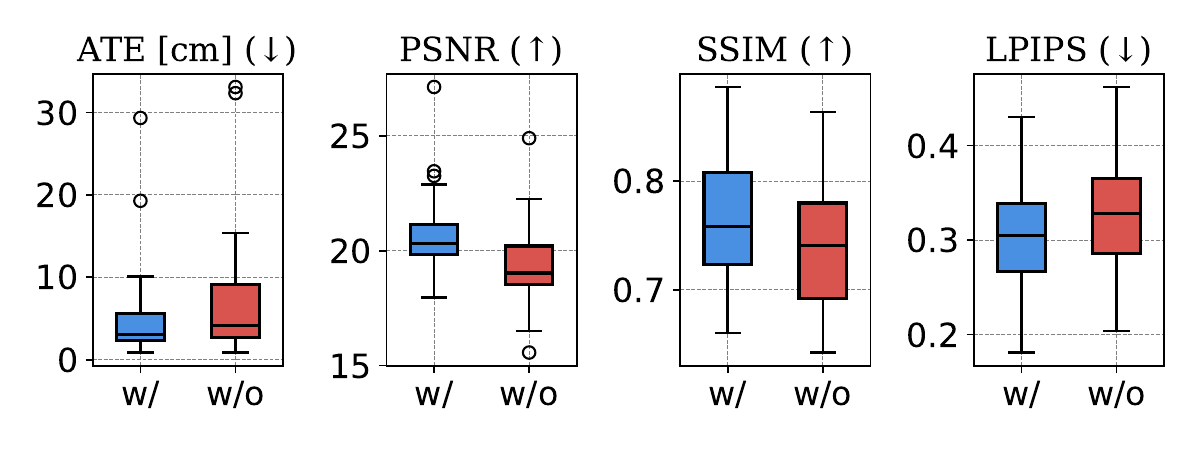}\vspace{-3mm}
    \phantomcaption  % Use \phantomcaption to avoid adding automatic (a) or (b) parentheses
    \textbf{(b)} \textit{7-Scenes}
\end{subfigure}\vspace{-2mm}
\caption{\textbf{Ablation study on collaborative pose optimization.}}
\label{fig:loop_closure_ablation}\vspace{-2mm}
\end{figure}

\vspace{0.5mm}
\noindent\textcolor{black}{
\textbf{Effect of  collaborative 3D map optimization via \textit{Inter-Agent Gaussian Consensus}.}
Fig.~\ref{fig:ma_sa_ablation_combined} shows that collaborative mapping optimization effectively improves performance on both synthetic and real-world datasets, with notable gains in the complex \textit{7-Scenes} dataset, particularly in mapping metrics like reduced LPIPS and higher PSNR and SSIM.}

\vspace{0.5mm}
\noindent\textcolor{black}{ \textbf{Effect of  collaborative pose optimization  via \textit{Inter-Agent Gaussian Consensus}.} Fig.~\ref{fig:loop_closure_ablation} illustrates that incorporating inter-agent consensus for global pose optimization significantly reduces tracking error and improves RGB image rendering quality from the reconstructed Gaussian Splat map, reflecting enhanced pose accuracy and visual fidelity. These gains are especially pronounced in the challenging \textit{7-Scenes} dataset, where dense and  ambiguous degraded observations intensify the importance of robust inter-agent consensus. }

%% file: sec/5_conclusion.tex
\section{Conclusion and Future Work}

We have presented MAC-Ego3D, a novel framework for real-time ego-motion estimation and photorealistic 3D reconstruction through multi-agent collaboration. MAC-Ego3D addresses key challenges in collaborative mapping by achieving low-latency pose tracking and high-fidelity 3D reconstruction. It leverages \textit{Multi-Agent Gaussian Consensus} to align and optimize 3D Gaussian splatting primitives both within and across agents, resulting in dense and coherent scene representations. Experimental results demonstrate that MAC-Ego3D outperforms existing methods in trajectory estimation accuracy, 3D map reconstruction fidelity, and runtime efficiency. We hope this work sheds light on multi-agent approaches for scalable  3D spatial intelligence.

\vspace{0.5mm}
\noindent\textbf{Limitations and future work}.
Scaling multi-agent collaborative SLAM to multiple rooms introduces the challenge of associating similar patterns across spaces, a direction beyond the scope of this work. Future extensions for MAC-Ego3D could include large-scale, unbounded outdoor 3D mapping~\cite{10342377}, enabling scalable spatial perception and high-fidelity world reconstruction. Additionally, as the 3D Gaussian splat representation can grow with scene scale, integrating effective Gaussian compression techniques~\cite{niedermayr2024compressed} will be essential for managing memory efficiency.

%% file: sec/X_suppl.tex
\setcounter{table}{0}  
\setcounter{figure}{0}  
\setcounter{section}{0}  
\renewcommand{\thetable}{\Alph{table}}
\renewcommand{\thefigure}{\Alph{figure}}
\renewcommand{\thesection}{\Alph{section}}

\section{More Qualitative Results}

We present a detailed qualitative comparison between our approach, MAC-Ego3D, and the previous SOTA multi-agent dense SLAM method, CP-SLAM~\cite{hu2024cp}. To showcase the generalization capabilities of MAC-Ego3D, we provide novel-view RGB and depth rendering results derived from its reconstructed 3D representations. Additionally, we visualize the Gaussian splat representation optimized by our multi-agent Gaussian consensus, showcasing the model’s ability to achieve photorealistic 3D reconstruction in real time.

To comprehensively demonstrate the spatial coherence of the rendered RGB-D outputs from the reconstructed 3D map of MAC-Ego3D and to compare our method with the previous state-of-the-art, CP-SLAM, we provide an immersive \textbf{\textcolor{mypurple}{video demo} in the supplementary material}. This visualization illustrates that \textbf{our MAC-Ego3D model generates high-fidelity, temporally consistent renderings of RGB-D frames} from the reconstructed 3D map, while CP-SLAM's predictions often suffer from missing regions, flickering effects, and spatial and temporal inconsistencies. Readers are encouraged to refer to the video for more insights.

\subsection{More RGB-D Rendering Results}

Figures~\ref{fig:rendering-comparison_supp1}, \ref{fig:rendering-comparison_supp2}, \ref{fig:rendering-comparison_supp3}, and \ref{fig:rendering-comparison_supp4} provide a detailed comparison of RGB and depth rendering results between MAC-Ego3D and CP-SLAM~\cite{hu2024cp}. Across both synthetic (\textit{Multi-agent Replica}) and real-world (\textit{7Scenes}~\cite{7scene}) datasets, MAC-Ego3D consistently delivers superior fidelity, continuity, and robustness in a variety of reconstruction scenarios.

\vspace{1mm}
\noindent\textbf{RGB rendering fidelity}. In RGB rendering, MAC-Ego3D generates continuous, photorealistic reconstructions with sharper texture details, as shown in Figures~\ref{fig:rendering-comparison_supp1}, \ref{fig:rendering-comparison_supp2}, and \ref{fig:rendering-comparison_supp3}. In contrast, CP-SLAM often struggles in high-frequency textured areas, producing fragmented and visually inconsistent outputs, which expose its limitations in capturing complex geometries during 3D reconstruction. MAC-Ego3D’s unified Gaussian splat representation, optimized through multi-agent Gaussian consensus, ensures smoother transitions and enhanced fidelity, particularly in regions with high-frequency details. Moreover, it effectively handles RGB-D video inputs captured in challenging environments featuring rich semantic objects and dynamic motion. Even in scenarios where CP-SLAM completely fails to reconstruct details, MAC-Ego3D demonstrates the ability to capture fine details with remarkable pose and RGB-D precision.

\vspace{1mm}
\noindent\textbf{Depth rendering fidelity}. Depth rendering comparisons, illustrated in Figure~\ref{fig:rendering-comparison_supp4}, underscore MAC-Ego3D's superiority in generating geometrically accurate and consistent depth maps. In contrast to CP-SLAM, which often exhibits abrupt discontinuities and fragmented geometry—especially in regions with sparse observations or intricate structures—MAC-Ego3D preserves depth continuity and structural integrity across diverse scenes. While minor blurring may occur in highly textured areas, the Gaussian splat representation optimized via multi-agent Gaussian consensus mitigates depth overfitting, a common issue with CP-SLAM, and enhances robustness across a variety of challenging scenarios.

\subsection{Novel-View RGB Rendering Results}
\begin{figure*}[t!]
\centering   %\setlength{\abovecaptionskip}{0.1cm}
\includegraphics[width=\textwidth]{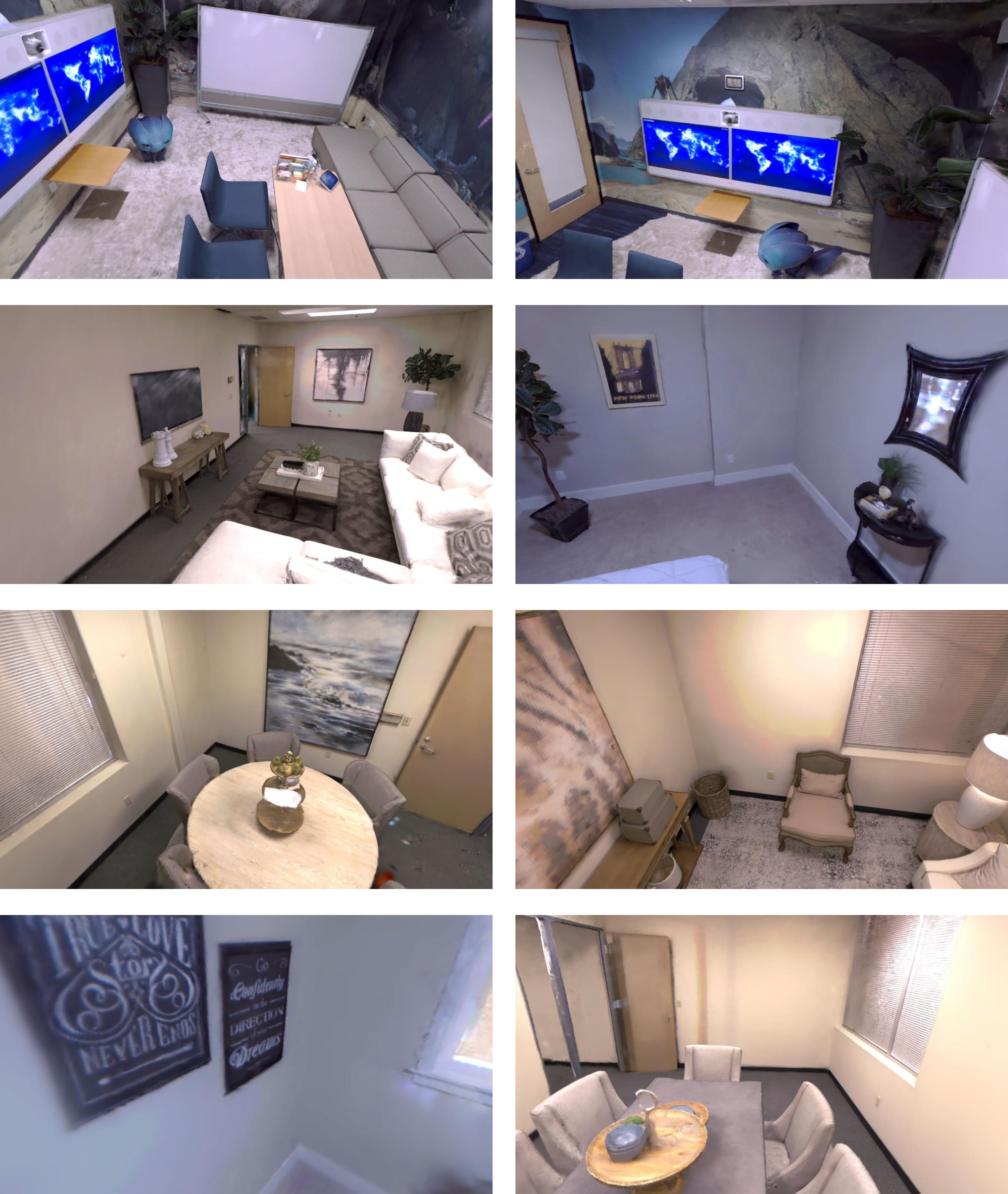}
\caption{Novel-view RGB image synthesis results via the proposed MAC-Ego3D model.}
\label{fig:nv_syn}
\end{figure*}

\begin{figure*}[t!]
\centering   %\setlength{\abovecaptionskip}{0.1cm}
\includegraphics[width=0.92\textwidth]{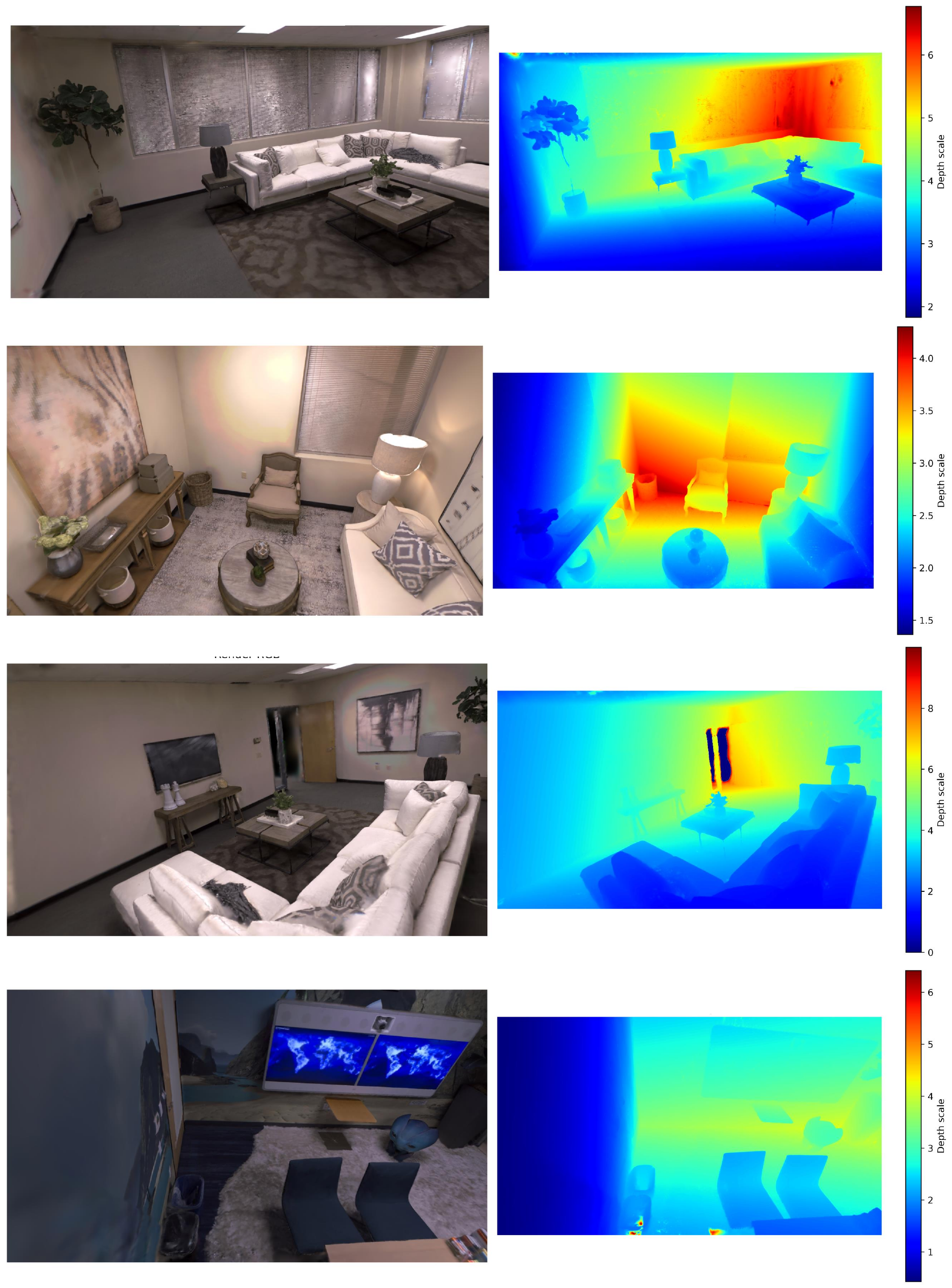}
\caption{Novel-view RGB and depth image synthesis results via the proposed MAC-Ego3D model.}
\label{fig:nv_syn1}\vspace{-4mm}
\end{figure*}

\begin{figure*}[t!]
\centering   %\setlength{\abovecaptionskip}{0.1cm}
\includegraphics[width=0.9\textwidth]{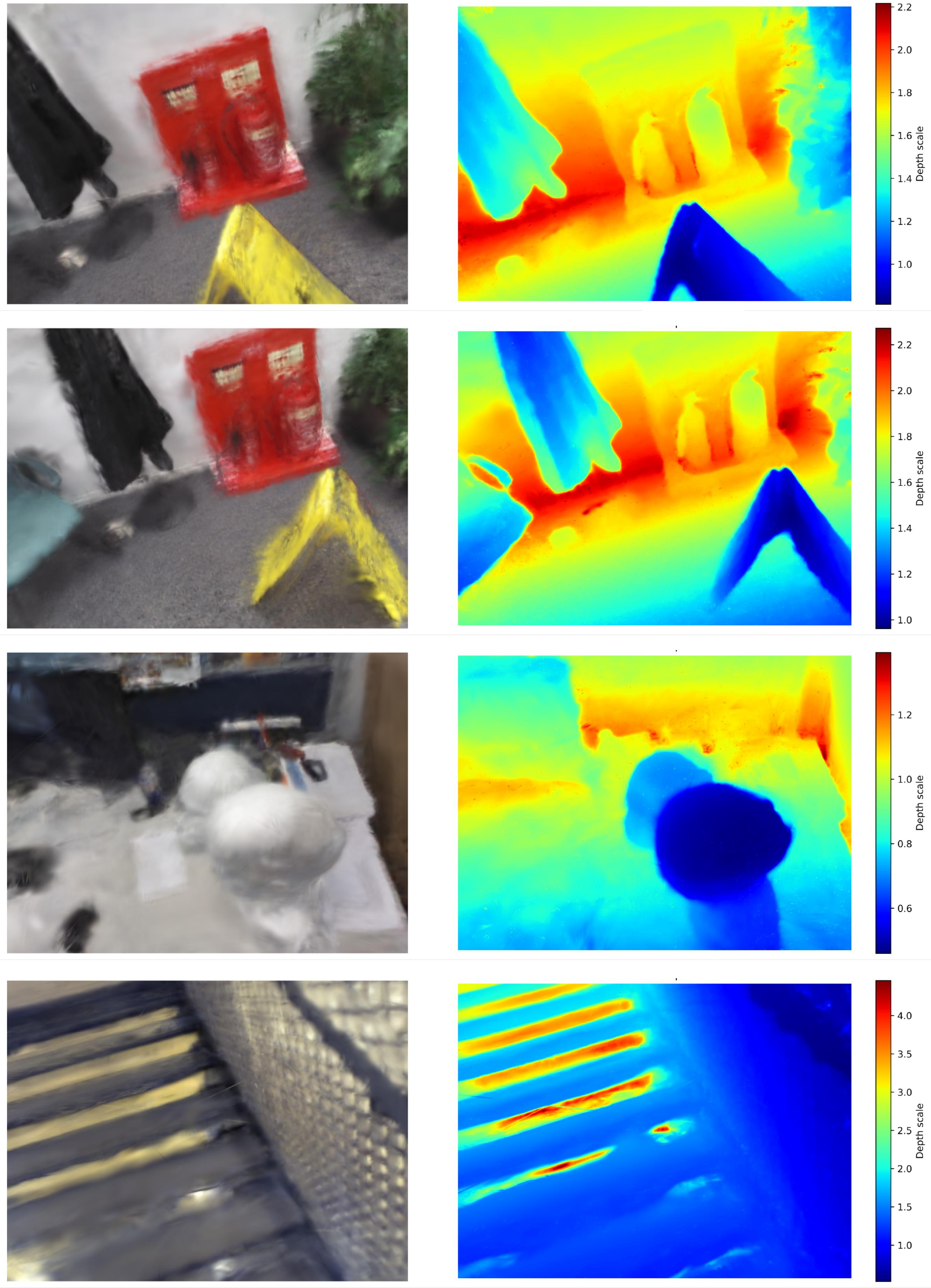}
\caption{Novel-view RGB and depth image synthesis results via the proposed MAC-Ego3D model.}
\label{fig:nv_syn2}\vspace{-4mm}
\end{figure*}

The novel-view RGB rendering capabilities of MAC-Ego3D are demonstrated in Figure~\ref{fig:nv_syn} and the left columns of Figures~\ref{fig:nv_syn1} and \ref{fig:nv_syn2}. MAC-Ego3D reliably synthesizes RGB images from unobserved viewpoints, reconstructing fine details with remarkable spatial coherence, particularly on the \textit{Multi-agent Replica} dataset, which benefits from clean depth observations. Its ability to seamlessly interpolate between observed views highlights the strength of the Gaussian splat-based representation in capturing and generalizing scene details beyond the training data. This capability establishes a new benchmark for generalization and photorealistic reconstruction in multi-agent SLAM systems.

\subsection{Novel-View Depth Rendering Results}
The right columns of Figures~\ref{fig:nv_syn1} and \ref{fig:nv_syn2} highlight MAC-Ego3D’s depth synthesis performance for novel viewpoints. The results demonstrate the model’s capability to accurately interpolate depth while maintaining geometric consistency across a range of challenging scenarios, including complex structures and sparsely observed regions. Notably, MAC-Ego3D achieves smooth and coherent transitions between viewpoints, underscoring its robustness in handling diverse and incomplete depth observations.

\subsection{3D Gaussian Splat Representation}

\begin{figure*}[t!]
\centering   %\setlength{\abovecaptionskip}{0.1cm}
\includegraphics[width=\textwidth]{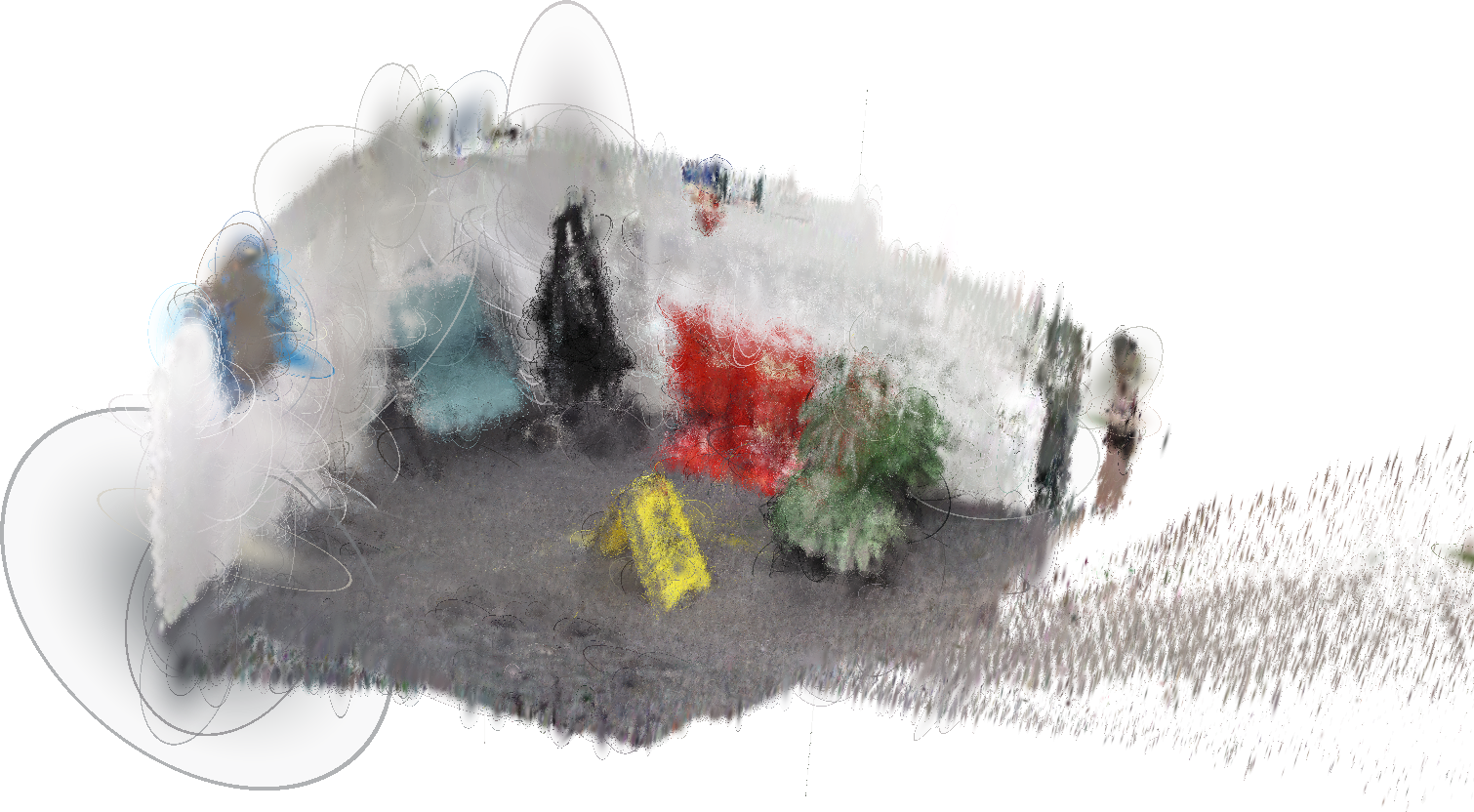}\vspace{1mm}
\includegraphics[width=\textwidth]{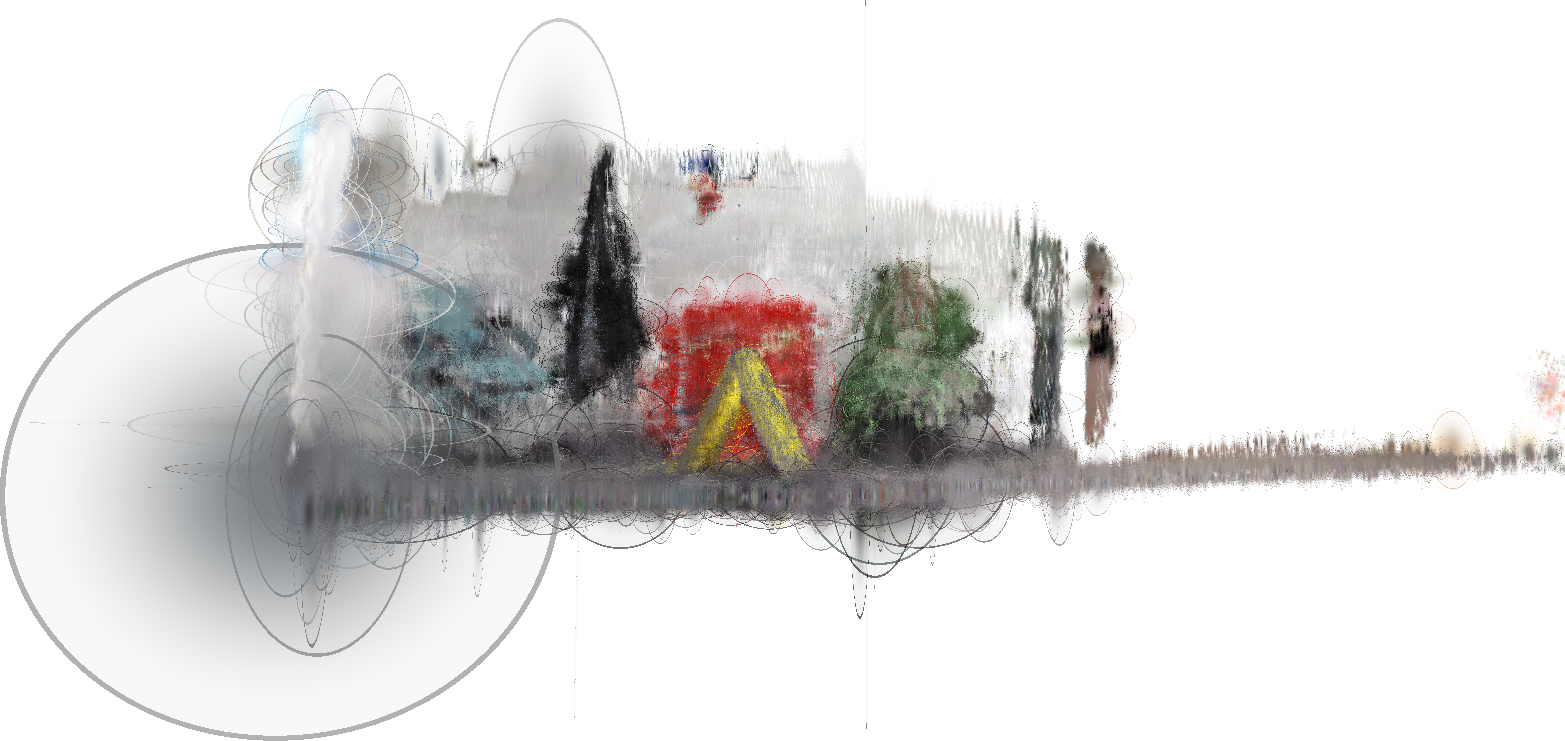}
\caption{Visualization of reconstructed 3D map with Gaussian splat representation.}
\label{fig:vis_gauss2}
\end{figure*}

\begin{figure*}[t!]
\centering   %\setlength{\abovecaptionskip}{0.1cm}
\includegraphics[width=\textwidth]{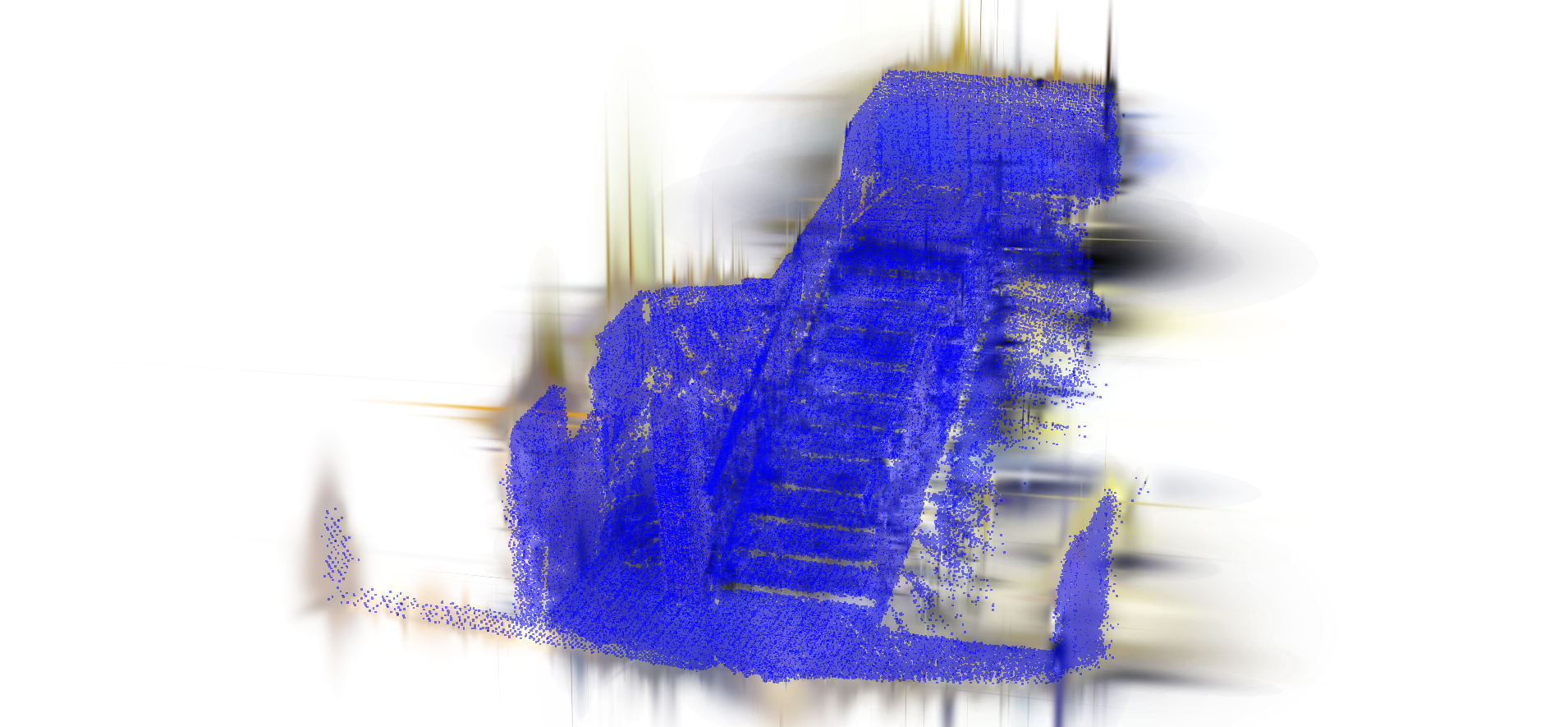}\vspace{1mm}
\includegraphics[width=\textwidth]{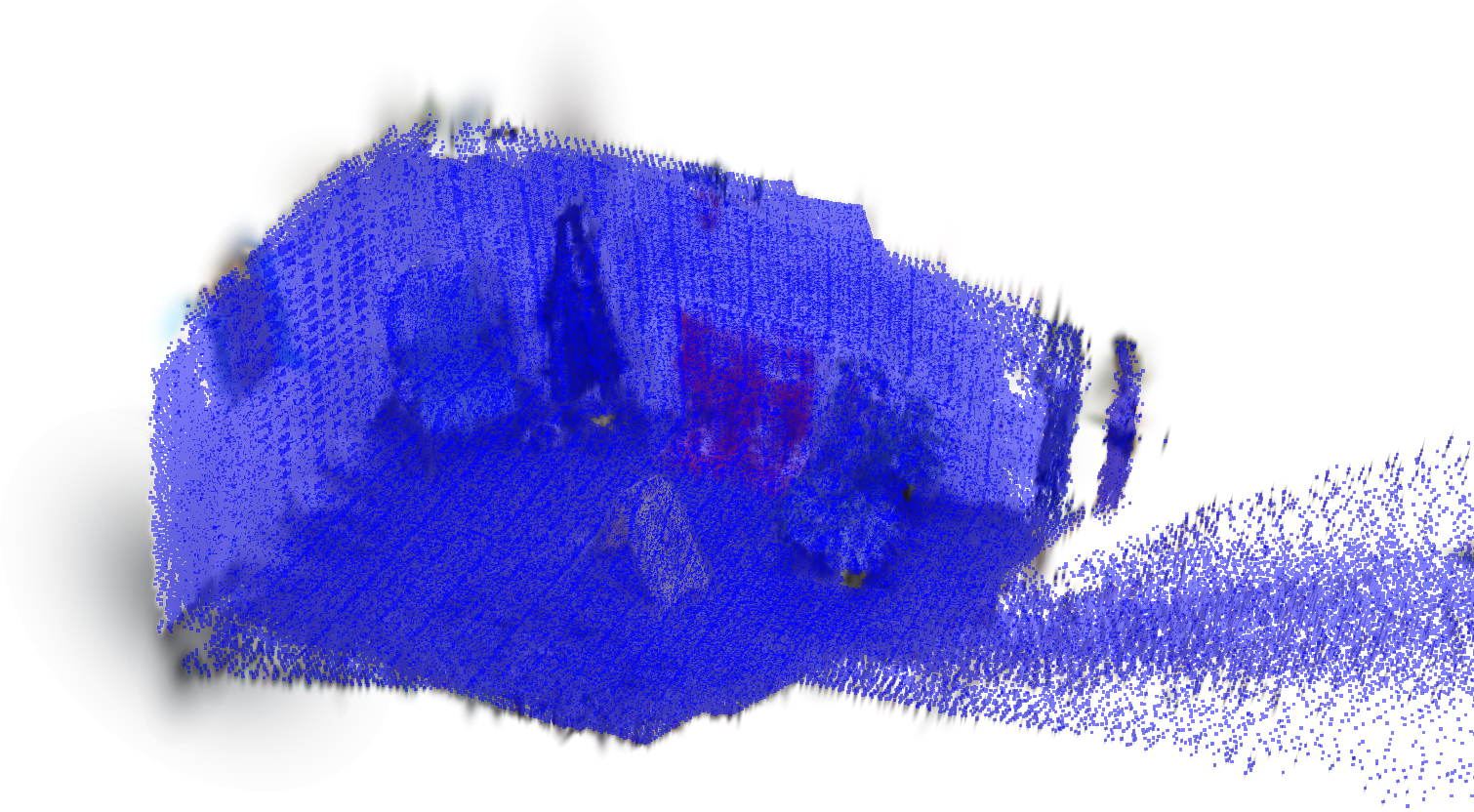}
\caption{Visualization of the coordinate centers (shown in blue points) of reconstructed 3D map with Gaussian splat representation.}
\label{fig:vis_gauss_center}
\end{figure*}

Figures~\ref{fig:vis_gauss1}, \ref{fig:vis_gauss2}, and \ref{fig:vis_gauss_center} illustrate the reconstructed 3D Gaussian splat maps, which encode both RGB color and geometry across various scenes. These visualizations highlight the spatial coherence and expressive power of the representation. By effectively balancing computational efficiency and visual accuracy, MAC-Ego3D enables real-time collaborative SLAM with high photorealistic fidelity.

\begin{figure*}[t!]
\centering   %\setlength{\abovecaptionskip}{0.1cm}
\includegraphics[width=0.95\textwidth]{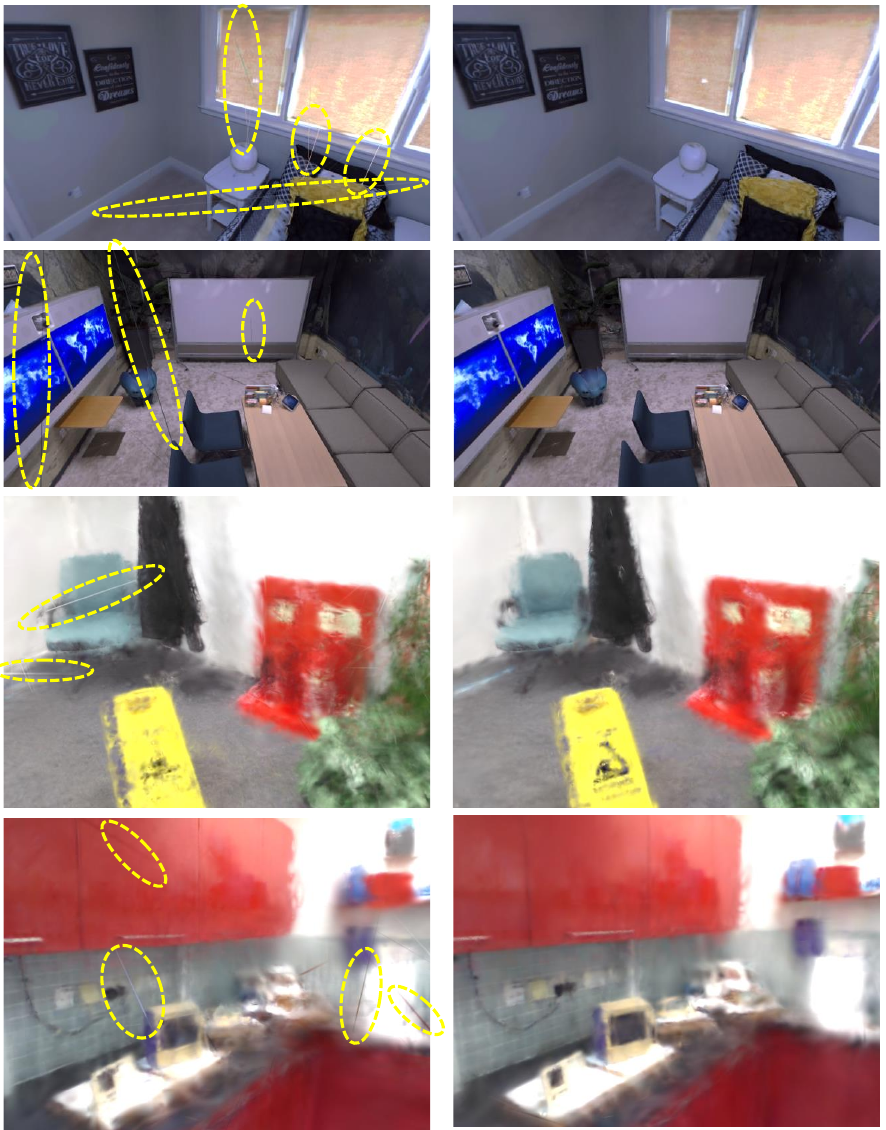}
\caption{\textbf{{Qualitative comparison of Gaussian splat pruning.}} The left column shows the raw Gaussian splats, while the right column presents the results after pruning. The pruning process effectively removes redundant or elongated splats, enhancing the rendering quality.}
\label{fig:gaussian_prune}\vspace{-4mm}
\end{figure*}

Figure~\ref{fig:gaussian_prune} visualizes the 3D Gaussian splat representation before and after the pruning process. The raw Gaussian splats (left panels in Figure~\ref{fig:gaussian_prune}) often include redundant and elongated primitives that adversely affect rendering quality. Our pruning method, outlined in Algorithm~\ref{alg:gaussian_pruning}, effectively removes these artifacts, resulting in a compact and visually accurate 3D representation (right panels in Figure~\ref{fig:gaussian_prune}). This pruning process enhances rendering fidelity by addressing over-redundancy and aligning splats more precisely.

\section{Hyperparameter Sensitivity Analysis}

\subsection{Similarity Threshold $\tau$ }
Figures~\ref{fig:sensitivity_tau_multiagent_replica} and~\ref{fig:sensitivity_tau_multiagent_7scenes} illustrate the impact of the similarity threshold $\tau$ on MAC-Ego3D’s performance for inter-agent overlap detection on the \textit{Multi-agent Replica} and \textit{7Scenes} datasets, respectively. The results indicate that MAC-Ego3D is robust to variations in $\tau$ within a local range of the default value.

\vspace{1mm}
\noindent\textbf{Trajectory estimation quality, measured by ATE.} On both datasets, larger $\tau$ values result in more aggressive overlap detection, slightly increasing trajectory errors due to reduced multi-agent collaborative map and pose optimization. Overly restrictive thresholds may miss valid overlaps, causing minor degradations in accuracy. Conversely, smaller $\tau$ values can increase false positives, occasionally leading to minor improvements in ATE. The default $\tau$ strikes a balance between these competing effects, ensuring stable performance.

\vspace{1mm}
\noindent\textbf{Mapping fidelity, measured by PSNR, SSIM, and LPIPS.} Metrics for mapping fidelity remain largely unaffected by variations in $\tau$, reflecting MAC-Ego3D’s ability to produce visually coherent outputs despite differences in inter-agent overlap quality. These findings highlight the model’s robustness, as small variations in $\tau$ do not significantly impact rendering quality or mapping accuracy.

\subsection{Communication Interval $T_{Comm}$}

Figures~\ref{fig:sensitivity_comm_multiagent_replica} and~\ref{fig:sensitivity_comm_multiagent_7scenes} illustrate the sensitivity of MAC-Ego3D’s performance to the communication interval $T_{Comm}$ on the \textit{Multi-agent Replica} and \textit{7Scenes} datasets. The results demonstrate that MAC-Ego3D maintains robust and consistent performance across a local range of the default $T_{Comm}$.

\vspace{1mm}
\noindent\textbf{Trajectory estimation quality, measured by ATE.} Shorter communication intervals (\textit{i.e.}, more frequent collaboration between agents) improve trajectory accuracy by enabling more frequent sharing and updates of the 3D map. This synchronization reduces inter-agent drift and ensures consistent alignment of Gaussian splats. In contrast, larger intervals, while more computationally efficient, introduce slight increases in trajectory error due to reduced inter-agent information sharing. The default $T_{Comm}$ achieves an optimal balance between computational efficiency and ATE performance, making MAC-Ego3D suitable for both real-time and high-accuracy ego-motion estimation and 3D reconstruction.

\vspace{1mm}
\noindent\textbf{Mapping fidelity, measured by PSNR, SSIM, and LPIPS.} Metrics such as PSNR and SSIM remain largely stable across different $T_{Comm}$ values, with marginal improvements observed at shorter intervals. Similarly, LPIPS remains consistently low, indicating high visual quality. These findings suggest that inter-agent communication frequency has minimal impact on rendering fidelity, further emphasizing the robustness of MAC-Ego3D’s multi-agent collaborative ego-motion estimation and 3D mapping framework.

\subsection{General Hyperparameter Robustness}

The sensitivity analyses confirm the robustness of MAC-Ego3D’s key hyperparameters, such as the similarity threshold $\tau$ and communication interval $T_{Comm}$, in multi-agent consensus optimization. Across a range of values, the model demonstrates stable performance without requiring precise tuning, underscoring its adaptability to varied cases.

Notably, the default hyperparameter settings strike a balance between computational efficiency and performance across both trajectory estimation and mapping fidelity metrics. The stability of $\tau$ ensures reliable overlap detection for multi-agent collaborative mapping, even in challenging scenarios with sparse or noisy observations. Similarly, $T_{Comm}$ balances the trade-off between frequent inter-agent synchronization for improved trajectory accuracy and computational efficiency required for real-time applications.

\section{More  Implementation Details}

\subsection{Gaussian Splat Pruning for Mapping}

\begin{algorithm}[t!]
\caption{\textbf{Gaussian Splat Pruning}}
\label{alg:gaussian_pruning}
\textbf{Input:} Opacity threshold $\tau_{o}$, scale threshold $\tau_s$, elongation threshold $\tau_e$, original  Gaussian set $\mathcal{M}$ \\
\textbf{Output:} Pruned Gaussian set $\mathcal{M}'$ 
\begin{algorithmic}[1]
    \State \textbf{Initialize:} $\mathcal{M}_{\text{prune}} \gets \emptyset$
    \State \textbf{Opacity Pruning:} $\mathcal{M}_{\text{opacity}} \gets \{\mathbf{G}_i \mid \lambda_i < \tau_{o}\}$
    \State $\mathcal{M}_{\text{prune}} \gets \mathcal{M}_{\text{prune}} \cup \mathcal{M}_{\text{opacity}}$
    \If {$\tau_s \neq \varnothing$}
        \State \textbf{Scale Pruning:} $\mathcal{M}_{\text{large}} \gets \{\mathbf{G}_i \mid \max(\mathbf{s}_i) > \tau_s\}$
        \State $\mathcal{M}_{\text{prune}} \gets \mathcal{M}_{\text{prune}} \cup \mathcal{M}_{\text{large}}$
        \State \textbf{Elongation Pruning:} $\mathcal{M}_{\text{elongated}} \gets \{\mathbf{G}_i \mid \max(\mathbf{s}_i) > \tau_e \cdot (\sum(\mathbf{s}_i) - \max(\mathbf{s}_i))\}$
        \State $\mathcal{M}_{\text{prune}} \gets \mathcal{M}_{\text{prune}} \cup \mathcal{M}_{\text{elongated}}$
    \EndIf
    \State \textbf{Remove Gaussians:} $\mathcal{M}' \leftarrow \mathcal{M} \setminus \mathcal{M}_{\text{prune}}$
    \State \textbf{Return:} $\mathcal{M}'$
\end{algorithmic}
\end{algorithm}

In the MAC-Ego3D framework, we enhance the 3D Gaussian map representation by introducing a pruning process to remove Gaussians that contribute minimally or negatively to the scene representation. Each Gaussian $\mathbf{G}_i$ is characterized by its mean position $\mathbf{x}_i$, covariance matrix $\mathbf{\Sigma}_i$, opacity $\lambda_i$, and scale dimensions $\mathbf{s}_i = \{s_{i1}, s_{i2}, s_{i3}\}$. The pruning process focuses on three key objectives: eliminating Gaussians with low opacity, excessively large Gaussians, and elongated Gaussians that can introduce artifacts in the final rendering.

\vspace{1mm}
\noindent\textbf{Opacity pruning.} The pruning process begins by evaluating the opacity $\lambda_i$ of each Gaussian. Gaussians with opacity below the threshold $\tau_{o}$ are considered negligible and removed:
\begin{equation}
    \mathcal{M}_{\text{opacity}} = \{\mathbf{G}_i \mid \lambda_i < \tau_{o}\}.
\end{equation}
By default, $\tau_{o}$ is set to 0.005.

\vspace{1mm}
\noindent\textbf{Scale pruning.}
To address excessively large Gaussians, the maximum scale dimension $\max(\mathbf{s}_i)$ is examined. Gaussians exceeding a fraction of the scale threshold $\tau_s$ are flagged for removal:
\begin{equation}
    \mathcal{M}_{\text{large}} = \{\mathbf{G}_i \mid \max(\mathbf{s}_i) > \tau_s\}.
\end{equation}
Here, $\tau_s$ is set to 0.25 by default.

\vspace{1mm}
\noindent\textbf{Elongation pruning.}
Rendering artifacts caused by elongated Gaussians are mitigated by introducing an elongation criterion. A Gaussian is classified as elongated if its largest scale dimension exceeds $\tau_e$ times the sum of the other two dimensions:
\begin{equation}
    \mathcal{M}_{\text{elongated}} = \{\mathbf{G}_i \mid \max(\mathbf{s}_i) > \tau_e \cdot (\sum(\mathbf{s}_i) - \max(\mathbf{s}_i))\}.
\end{equation}
In our implementation, $\tau_e$ is set to 10 to target needle-like artifacts. As illustrated in Fig.~\ref{fig:gaussian_prune}, this step effectively removes elongated Gaussians, significantly improving the fidelity of the rendered images.

\vspace{1mm}
\noindent\textbf{Combined pruning.}
The final pruning mask combines all three criteria:
\begin{equation}
    \mathcal{M}_{\text{prune}} = \mathcal{M}_{\text{opacity}} \cup \mathcal{M}_{\text{large}} \cup \mathcal{M}_{\text{elongated}}.
\end{equation}
The flagged Gaussians are removed from the map:
\begin{equation}
    \mathcal{M}' \leftarrow \mathcal{M} \setminus \mathcal{M}_{\text{prune}}.
\end{equation}

This pruning process ensures a compact and efficient 3D map representation while maintaining high visual fidelity in the rendered scenes. The full procedure is detailed in Algorithm~\ref{alg:gaussian_pruning}.

\subsection{Keyframe Sampling for Pose Tracking}

Building on prior work in Gaussian splatting SLAM~\cite{ha2025rgbd,orbslam}, we employ an adaptive keyframe sampling strategy to enhance pose tracking efficiency. This method leverages geometric correspondences derived from G-ICP to selectively sample keyframes based on a correspondence threshold, ensuring consistent tracking performance while maintaining a uniform Gaussian density within the map. 

Unlike the selective approach used for tracking, every one of ten frames are utilized during mapping to fully exploit the RGB-D observations collected across all agents. This dual strategy enhances robustness by incorporating only non-overlapping Gaussians during pose tracking, thereby reducing accumulated tracking errors. It also achieves an optimal balance between tracking efficiency and mapping completeness, maximizing overall reconstruction accuracy. The geometric correspondence threshold for triggering pose tracking follows the setup described in~\cite{ha2025rgbd}.

\subsection{Construction of Testing Cases on the Real-World \textit{7Scenes} Dataset}

To evaluate the performance of our method under real-world conditions, we construct testing cases using the \textit{7Scenes} dataset, which features diverse indoor environments with complex visual and geometric structures. The testing setup is carefully designed to benchmark our approach through comparisons with prior methods and ablation studies, assessing its adaptability and robustness.

\vspace{1mm}
\noindent \textbf{Benchmarking performance against CP-SLAM.}  
Since CP-SLAM supports only two-agent collaboration, we configure testing cases by selecting two sequences from each scene in the \textit{7Scenes} dataset. For most scenes, sequences \texttt{seq-01} and \texttt{seq-02} are used, as they provide sufficient spatial overlap to enable collaboration. However, for the \textit{redkitchen} scene, we use sequences \texttt{seq-01} and \texttt{seq-03}, as these sequences cover complementary areas, ensuring diverse and representative results. This setup allows for fair and consistent performance comparisons with CP-SLAM.

\vspace{1mm}
\noindent \textbf{Testing cases for multi-agent collaboration in ablation studies.}  
To evaluate our method’s flexibility in supporting multi-agent collaboration with varying numbers of agents, we select sequences from the same scene with overlapping spatial regions to simulate realistic collaboration scenarios. Sequences are grouped based on their serial numbers to ensure even distribution and adequate overlap for inter-agent collaboration. Further, we select the best-performing sequence groups to conduct additional experiments, providing deeper insights into our method’s adaptability. The configurations for all testing cases are detailed in Table~\ref{tab:experimental_results}.

\begin{table}[t!]
\centering
 \centering \setlength{\tabcolsep}{5mm}
\setlength{\abovecaptionskip}{0.1cm}
\caption{Multi-agent testing case configurations for the ablation studies on the \textit{\textit{\textit{7Scenes}}} dataset.}
\resizebox{\linewidth}{!}{
\begin{tabular}{lcl}
\toprule
\textbf{Scene} & \textbf{Testing Case} & \textbf{Sequences} \\
\midrule
\multirow{3}{*}{\textbf{Chess}} 
& 1.1 & 01, 02, 03 \\ 
& 1.2 & 04, 05, 06 \\ 
& 1.3 & 01, 03, 05 \\ 
\midrule
\textbf{Fire} 
& 2.1 & 01, 02, 03, 04 \\ 
\midrule
\textbf{Heads} 
& 3.1 & 01, 02 \\ 
\midrule
\multirow{4}{*}{\textbf{Office}} 
& 4.1 & 01, 02, 03 \\ 
& 4.2 & 04, 05, 06 \\ 
& 4.3 & 07, 08, 09, 10 \\ 
& 4.4 & 01, 02, 04 \\ 
\midrule
\multirow{4}{*}{\textbf{Redkitchen}} 
& 5.1 & 01, 02, 03, 04 \\ 
& 5.2 & 05, 06, 07, 08 \\ 
& 5.3 & 11, 12, 13, 14 \\ 
& 5.4 & 01, 03, 04 \\ 
\midrule
\multirow{3}{*}{\textbf{Stairs}} 
& 6.1 & 01, 02, 03 \\ 
& 6.2 & 04, 05, 06 \\ 
& 6.3 & 01, 02, 04 \\ 
\bottomrule
\end{tabular}}
\label{tab:experimental_results}
\end{table}

\section{More Details on Evaluation Metrics}

Following prior works in dense Neural SLAM~\cite{hu2024cp,keetha2024splatam,ha2025rgbd}, we evaluate the proposed system using metrics including Absolute Trajectory Error (ATE), Peak Signal-to-Noise Ratio (PSNR), Structural Similarity Index Measure (SSIM), Learned Perceptual Image Patch Similarity (LPIPS), and Depth L1 Error. These metrics assess the fidelity of predictions relative to clean ground truth under two input conditions: (i) clean RGB-D observations $\mathbf{Z}_{1:t}^{a_i} = \{ (\mathbf{I}_k^{a_i}, \mathbf{D}_k^{a_i}) \}_{k=1}^t$ for agent $a_i$, 
 under the standard evaluation settings and (ii) partially perturbed observations in either the RGB $\mathbf{I}_k^{a_i}$ or depth $\mathbf{D}_k^{a_i}$ modalities  under the robustness evaluation settings. 

Specifically, ATE evaluates the spatial alignment between the estimated trajectory $\mathbf{T}_{1:t}^{a_i}$ and the ground truth trajectory $\mathbf{T}_{1:t}^\text{GT}$ as the root mean square error:
\begin{equation}
\text{ATE} = \sqrt{\frac{1}{N} \sum_{k=1}^{N} \|\mathbf{T}_k^{a_i} - \mathbf{T}_k^\text{GT}\|^2}.
\end{equation}

\noindent PSNR quantifies the pixel-level fidelity of the reconstructed RGB image $\mathbf{I}_k^{a_i}$ relative to the ground truth $\mathbf{I}_k^\text{GT}$:
\begin{equation}
\text{PSNR} = 10 \cdot \log_{10}\left(\frac{\text{MAX}^2}{\text{MSE}}\right)  [dB],
\end{equation}
where $\text{MAX}$ is the maximum pixel intensity and $\text{MSE}$ is the mean squared error. 

SSIM evaluates structural similarity, incorporating luminance, contrast, and structural information:
\begin{equation}
\text{SSIM} = \frac{(2\mu_x\mu_y + C_1)(2\sigma_{xy} + C_2)}{(\mu_x^2 + \mu_y^2 + C_1)(\sigma_x^2 + \sigma_y^2 + C_2)},
\end{equation}
where $\mu_x$, $\mu_y$, $\sigma_x$, and $\sigma_y$ denote the means and variances of $\mathbf{I}_k^{a_i}$ and $\mathbf{I}_k^\text{GT}$, while $\sigma_{xy}$ represents their covariance.

LPIPS~\cite{zhang2018unreasonable} measures perceptual similarity between the predicted RGB image $\mathbf{I}_k^{a_i}$ and the ground truth $\mathbf{I}_k^\text{GT}$, based on feature-level differences from a pre-trained neural network. The LPIPS score is:
\begin{equation}
\text{LPIPS} = \sum_{l} w_l \cdot \|\phi_l(\mathbf{I}_k^{a_i}) - \phi_l(\mathbf{I}_k^\text{GT})\|_2^2,
\end{equation}
where $\phi_l(\cdot)$ denotes feature maps from the $l$-th layer of a pre-trained network, with $w_l$ as learned layer weights.

Finally, Depth L1 Error quantifies the geometric accuracy of the predicted depth $\mathbf{D}_k^{a_i}$ compared to the ground truth depth $\mathbf{D}_k^\text{GT}$:
\begin{equation}
\text{Depth L1} = \frac{1}{N} \sum_{k=1}^{N} \|\mathbf{D}_k^{a_i} - \mathbf{D}_k^\text{GT}\|.
\end{equation}

{
    \small
    \bibliographystyle{ieeenat_fullname}
    \bibliography{main}
}
\clearpage

\begin{figure*}[t!]
\centering   \setlength{\abovecaptionskip}{0.1cm}
\includegraphics[width=\textwidth]{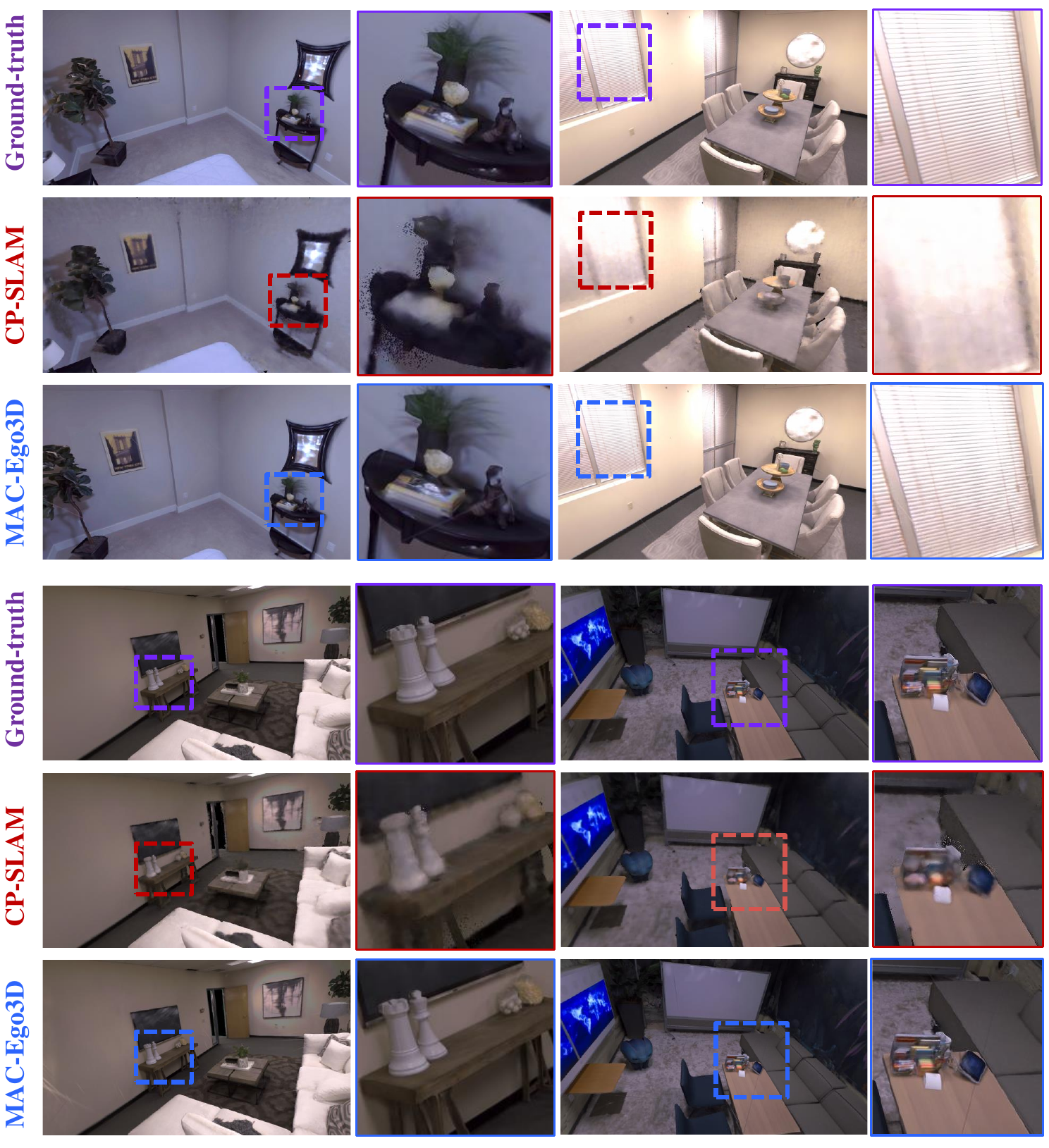}
\caption{\textbf{Qualitative RGB image rendering quality comparison between multi-agent SLAM models with dense reconstruction capability}, \textit{i.e.}, CP-SLAM and our MAC-Ego3D, on the \textit{\textit{\textit{Multi-agent Replica}}} dataset.}
\label{fig:rendering-comparison_supp1}
\end{figure*}

\begin{figure*}[t!]
\centering   \setlength{\abovecaptionskip}{0.1cm}
\includegraphics[width=\textwidth]{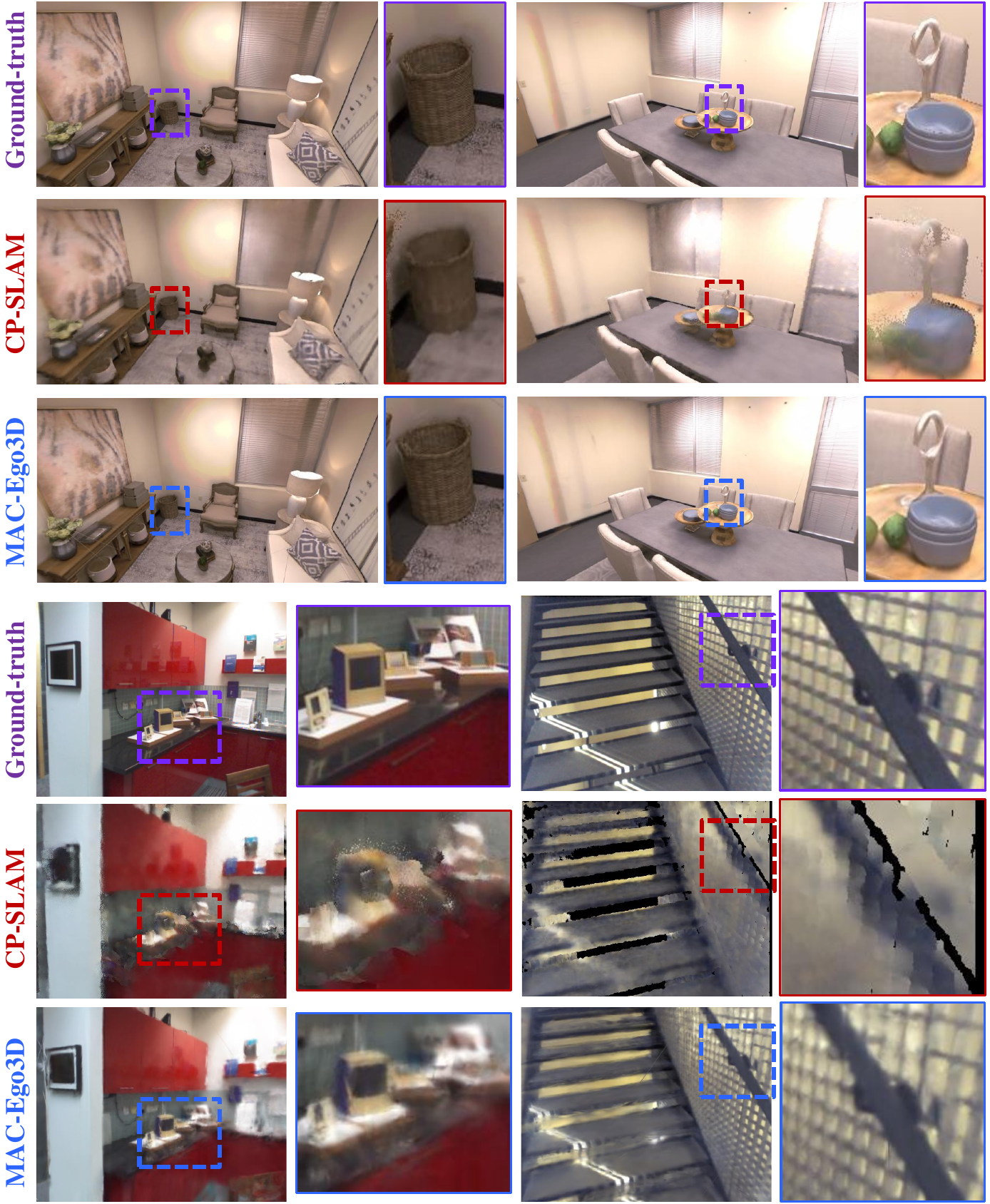}
\caption{\textbf{Qualitative RGB image rendering quality comparison between multi-agent SLAM models with dense reconstruction capability}, \textit{i.e.}, CP-SLAM and our MAC-Ego3D, on the \textit{\textit{\textit{Multi-agent Replica}}} and  \textit{7-Scenes}  
 dataset.}
\label{fig:rendering-comparison_supp2}
\end{figure*}

\begin{figure*}[t!]
\centering   \setlength{\abovecaptionskip}{0.1cm}
\includegraphics[width=\textwidth]{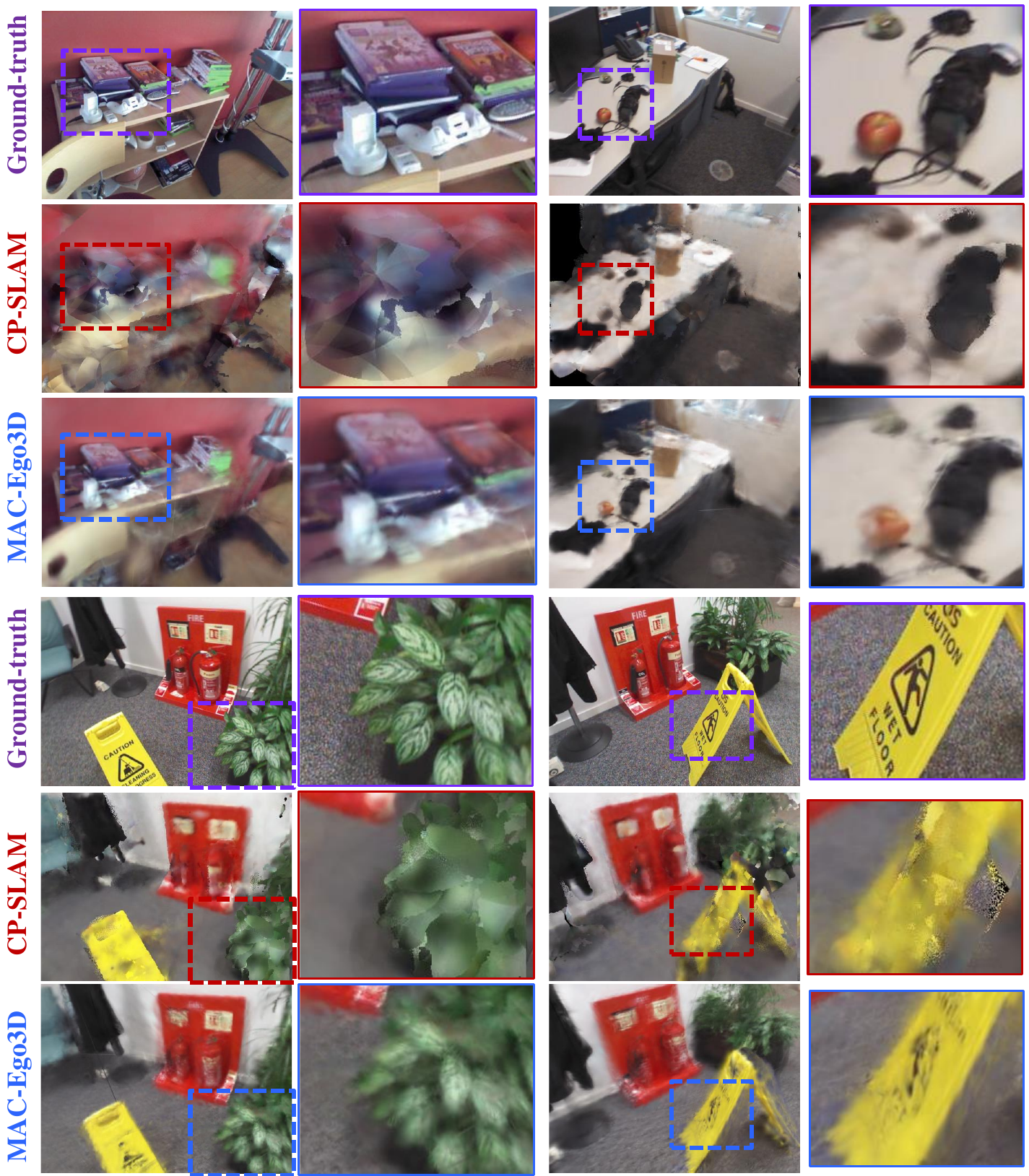}
\caption{\textbf{Qualitative RGB image rendering quality comparison between multi-agent SLAM models with dense reconstruction capability}, \textit{i.e.}, CP-SLAM and our MAC-Ego3D, on the  \textit{7-Scenes}  dataset.}
\label{fig:rendering-comparison_supp3}
\end{figure*}

\begin{figure*}[t!]
\centering   \setlength{\abovecaptionskip}{0.1cm}
\includegraphics[width=\textwidth]{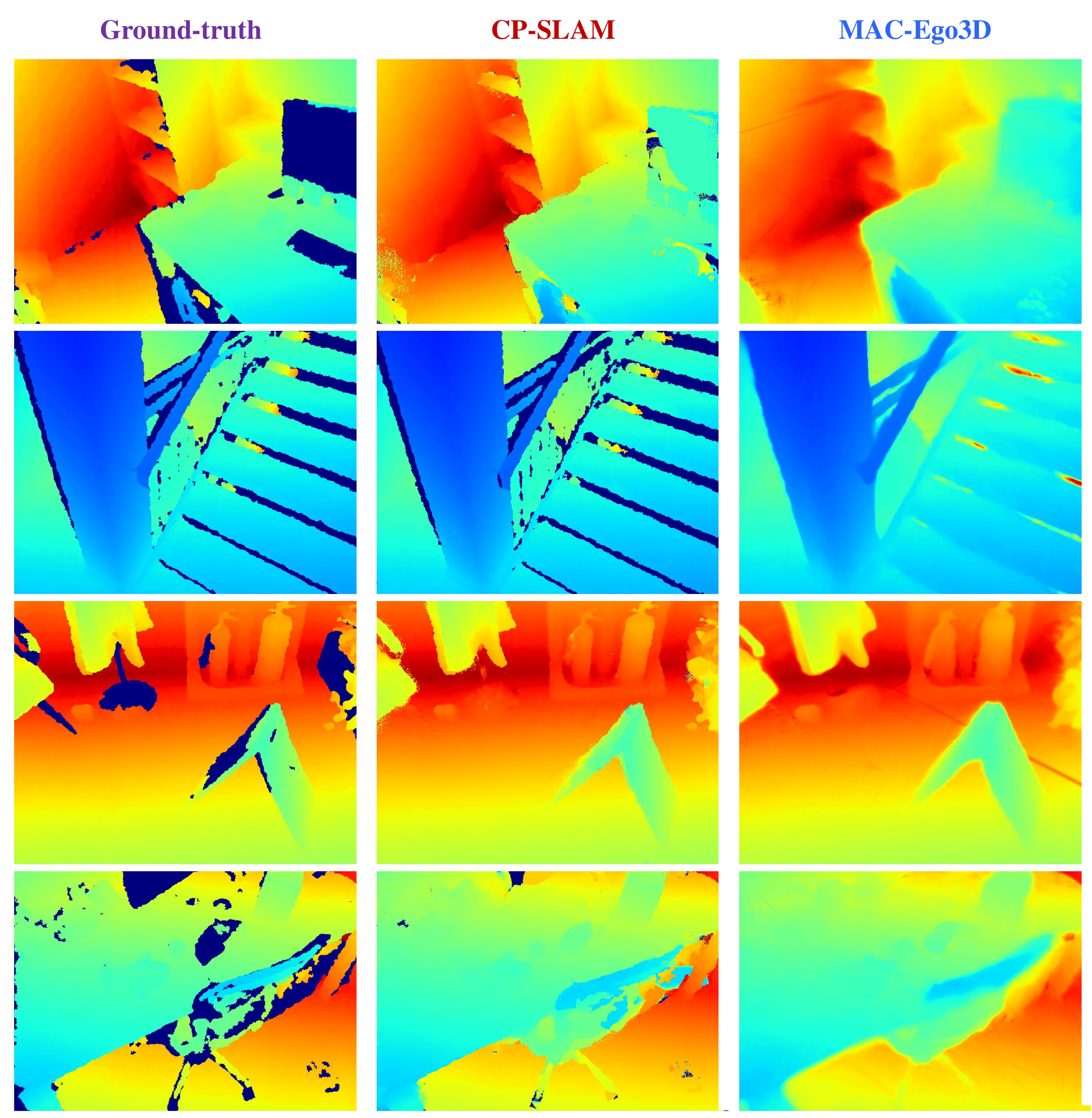}
\caption{ \textbf{Qualitative depth image rendering quality comparison between multi-agent SLAM models with dense reconstruction capability}, \textit{i.e.}, CP-SLAM and our MAC-Ego3D, on \textit{\textit{\textit{Multi-agent Replica}}} (\textbf{Left}) and \textit{7-Scenes} (\textbf{Right}) datasets.}
\label{fig:rendering-comparison_supp4}
\end{figure*}

\begin{figure*}[t!]
\centering   \setlength{\abovecaptionskip}{0.1cm}
\includegraphics[width=0.55\textwidth]{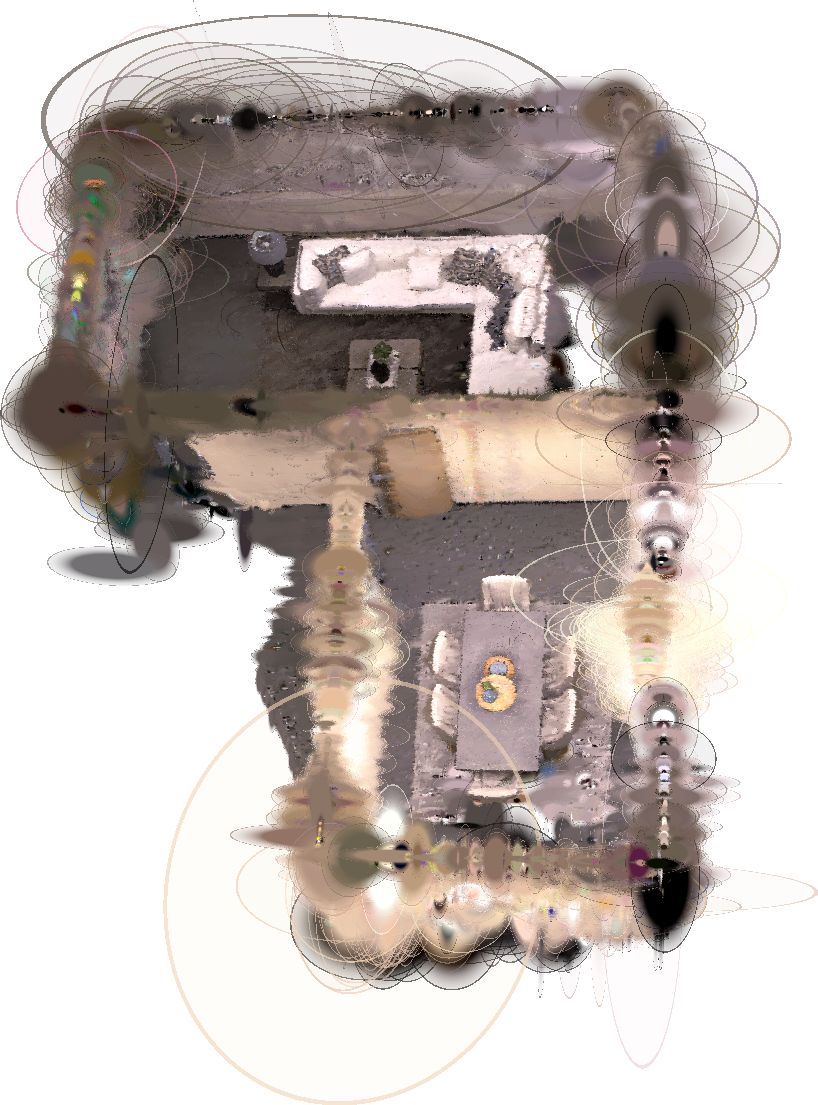}
\includegraphics[width=\textwidth]{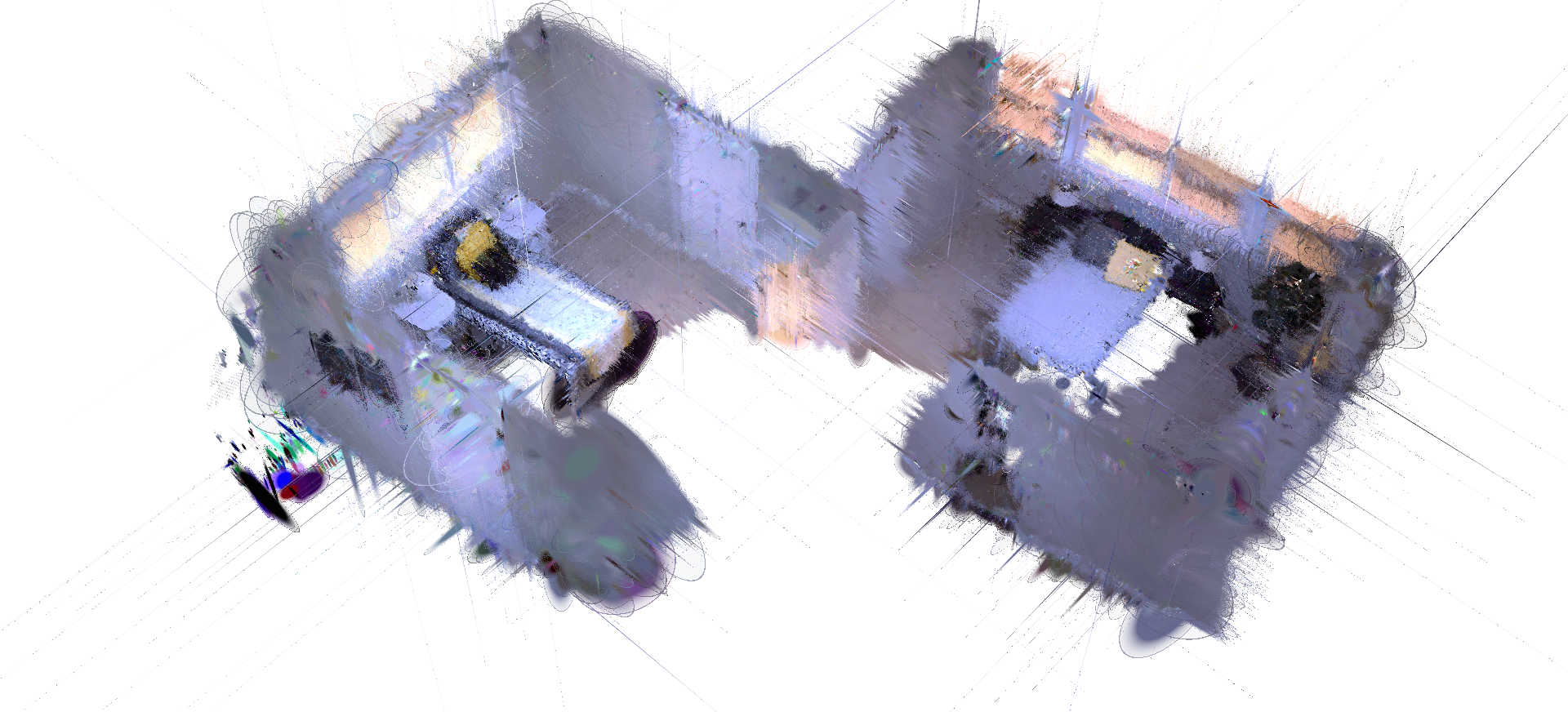}
\caption{Visualization of reconstructed 3D map with Gaussian splat representation.}
\label{fig:vis_gauss1}
\end{figure*}

\clearpage
\clearpage

\begin{figure}[t!]
\centering \setlength{\abovecaptionskip}{0.2cm}
\includegraphics[width=0.49\textwidth]{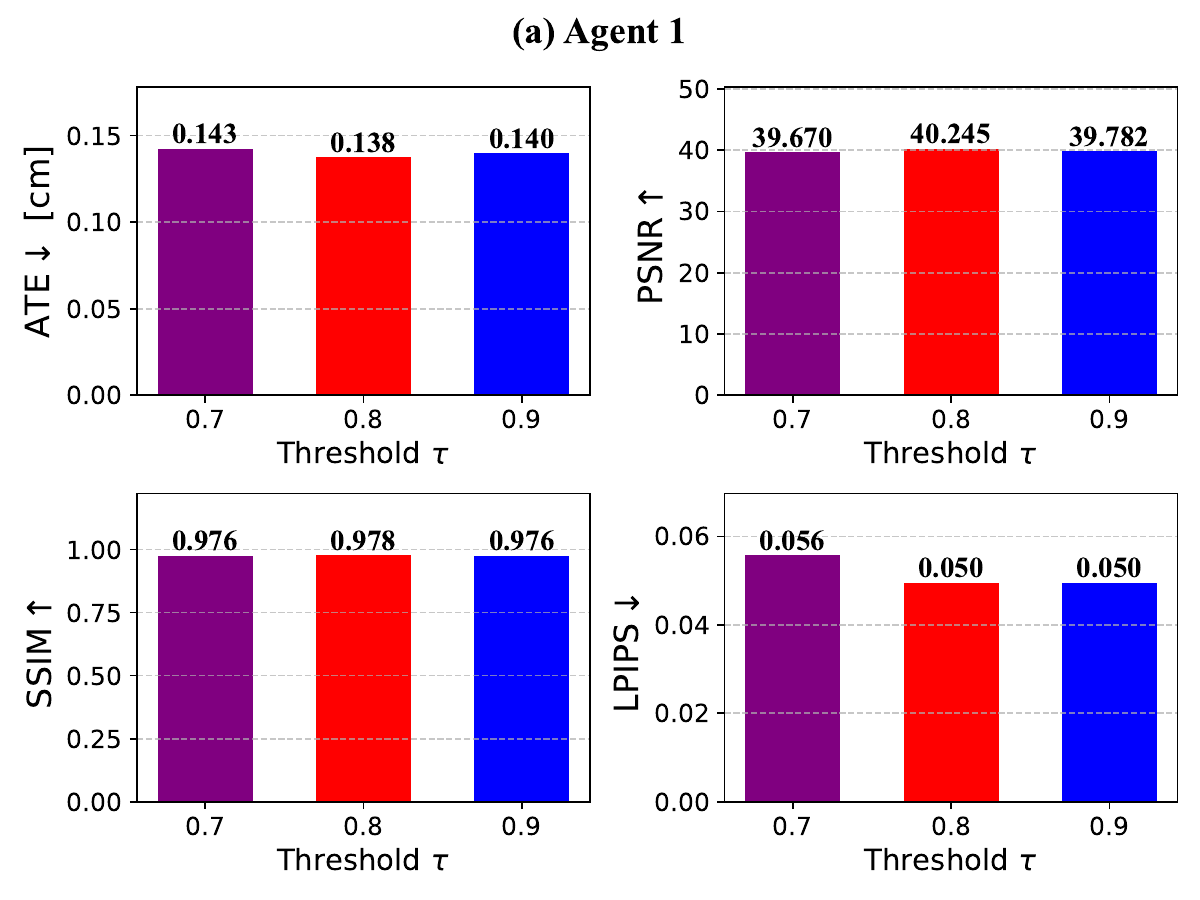}
\includegraphics[width=0.49\textwidth]{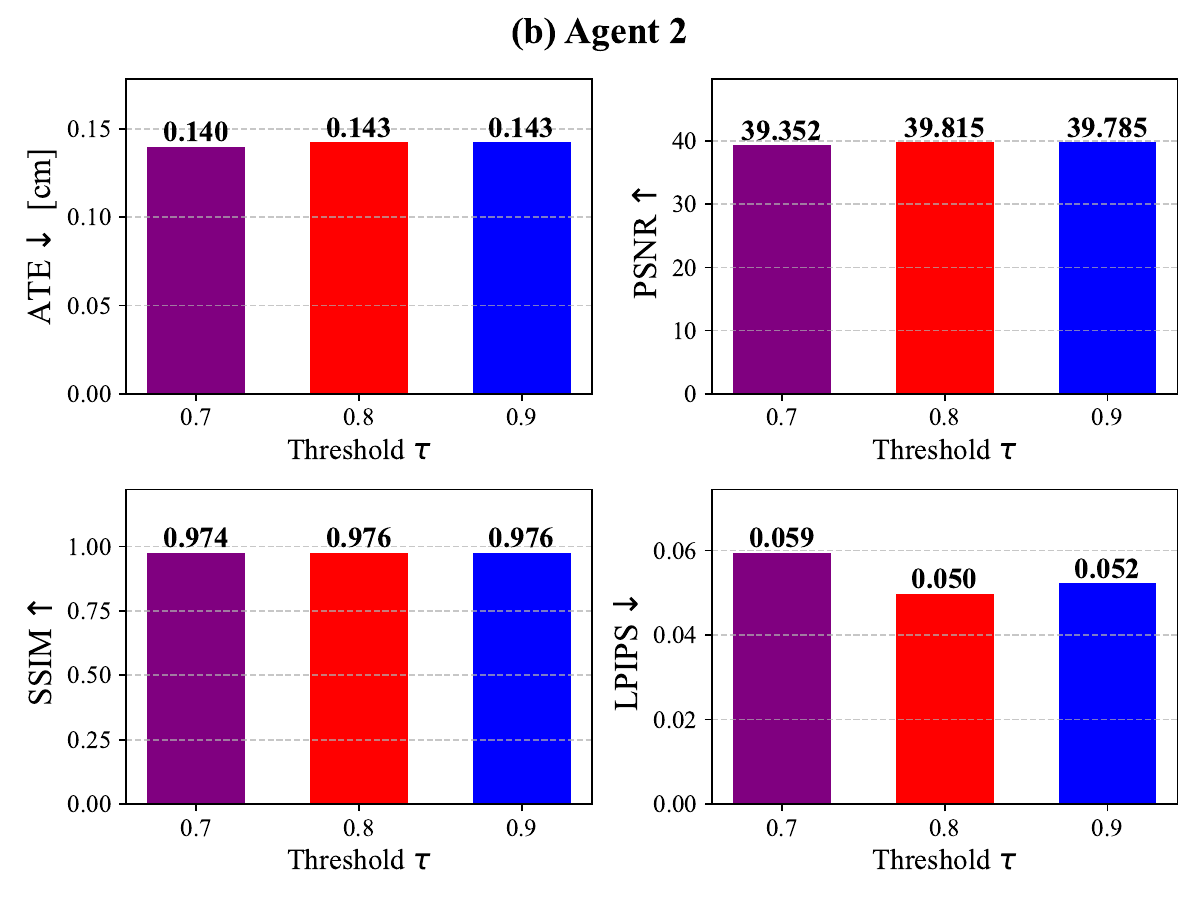}
\includegraphics[width=0.49\textwidth]{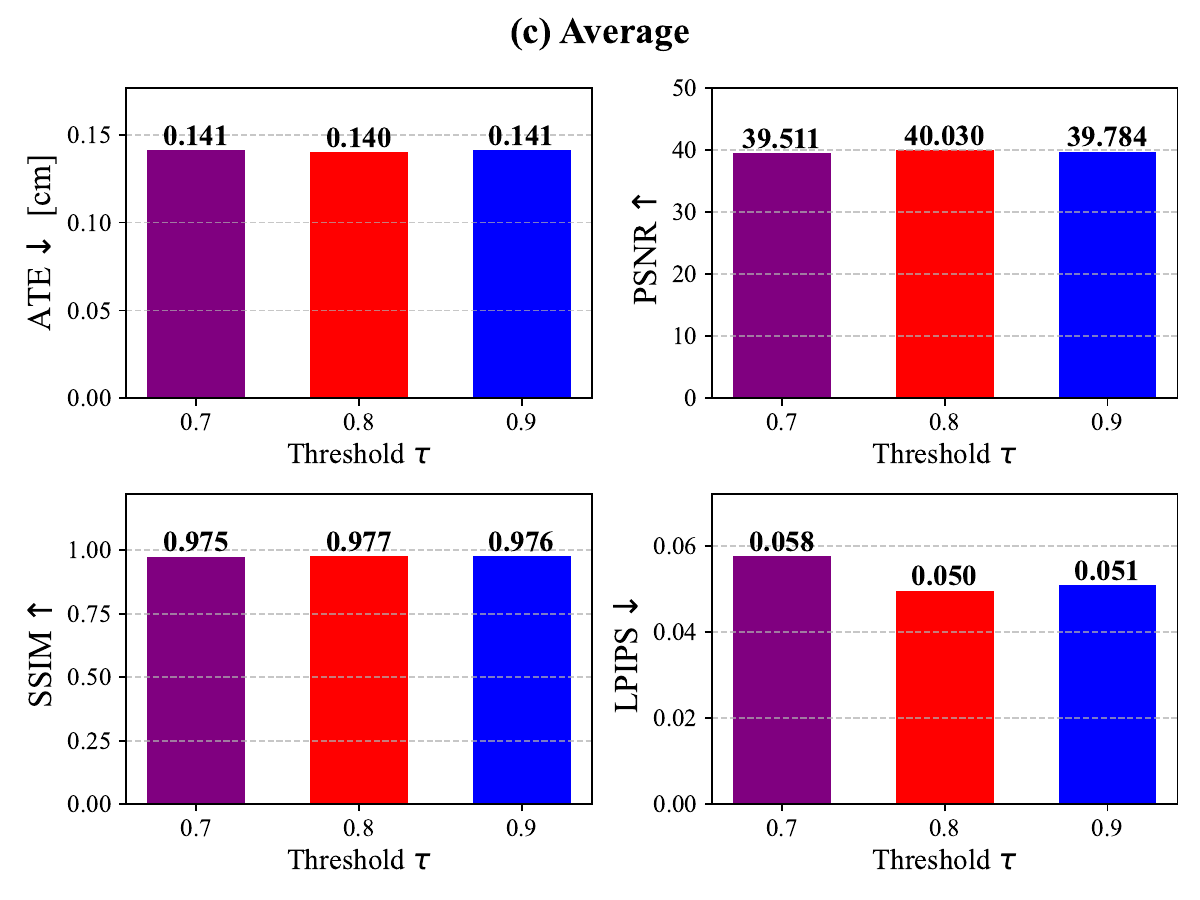}
\caption{\textbf{Hyperparameter sensitivity  analysis of the threshold $\tau$ for inter-agent overlap detection on \textit{\textit{\textit{Multi-agent Replica}}}. } } 
\label{fig:sensitivity_tau_multiagent_replica}\vspace{-2mm}
\end{figure}

\begin{figure}[t!]
\centering \setlength{\abovecaptionskip}{0.2cm}
\includegraphics[width=0.49\textwidth]{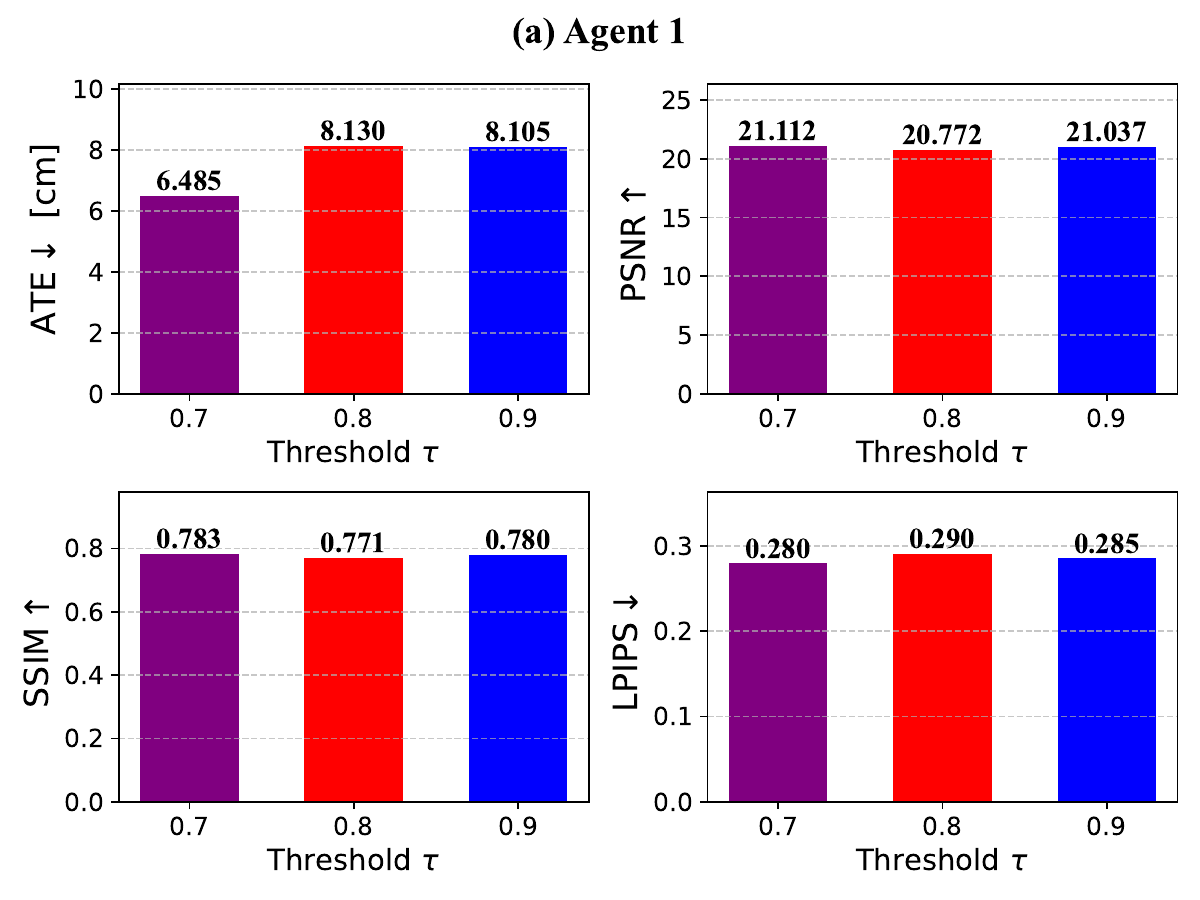}
\includegraphics[width=0.49\textwidth]{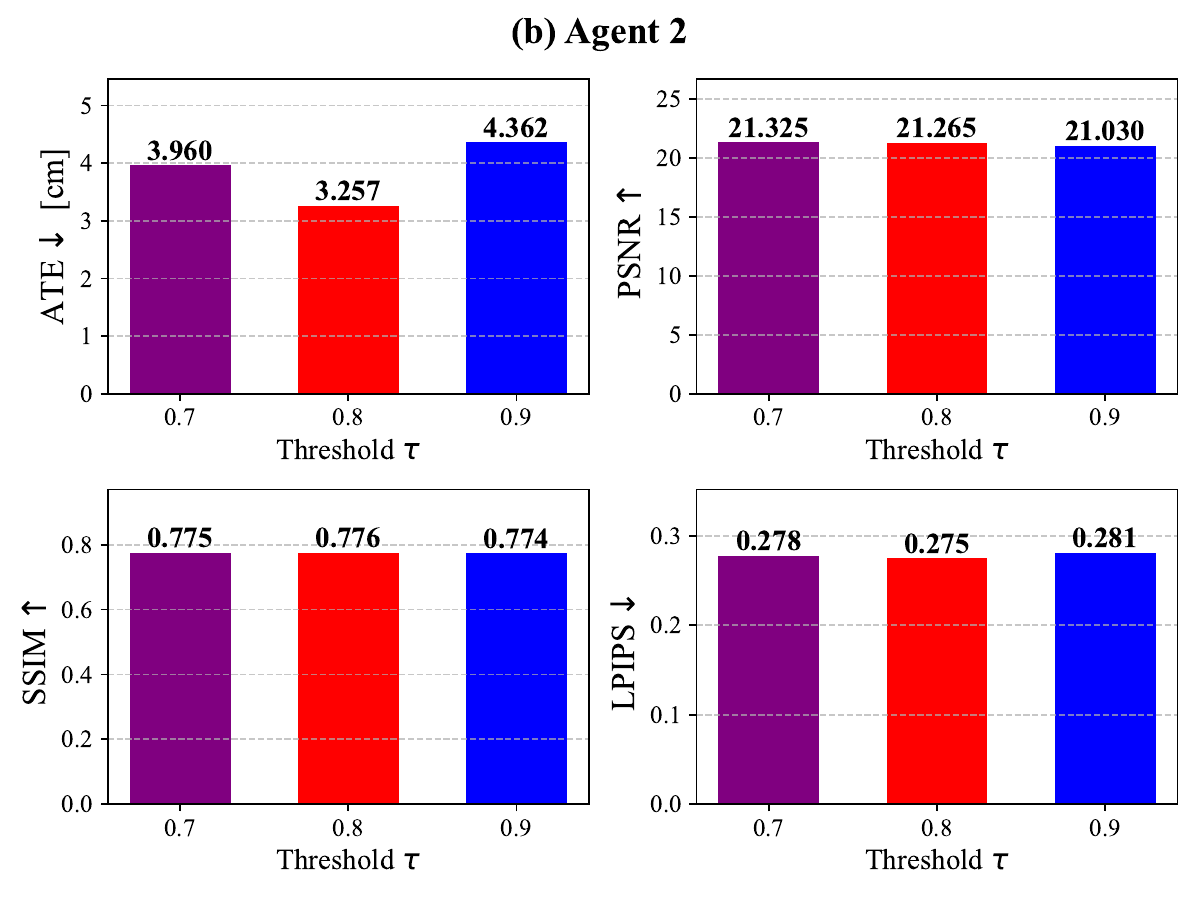}
\includegraphics[width=0.49\textwidth]{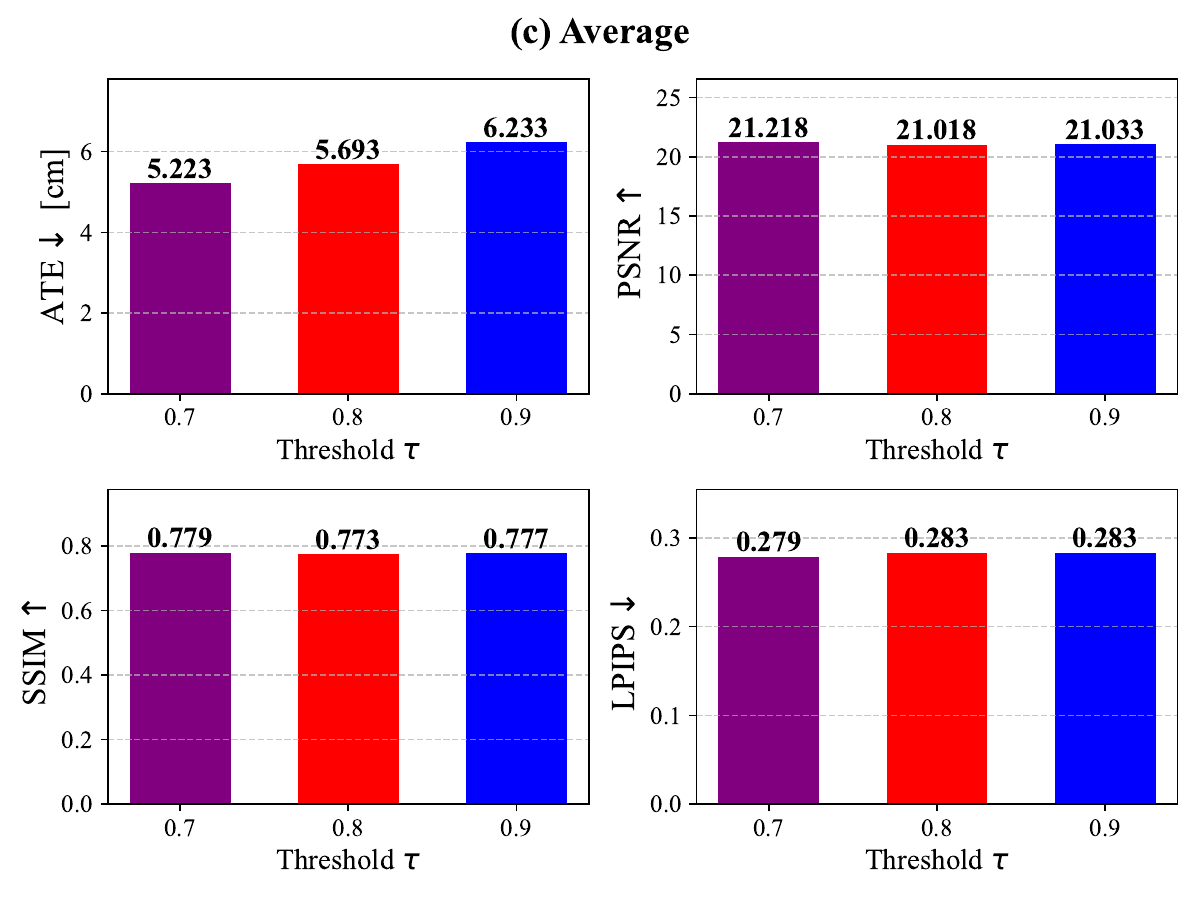}
\caption{\textbf{Hyperparameter sensitivity  analysis of the similarity threshold $\tau$ for inter-agent overlap detection on \textit{\textit{\textit{7Scenes}}}. } } 
\label{fig:sensitivity_tau_multiagent_7scenes}\vspace{-2mm}
\end{figure}

\begin{figure}[t!]
\centering \setlength{\abovecaptionskip}{0.2cm}
\includegraphics[width=0.49\textwidth]{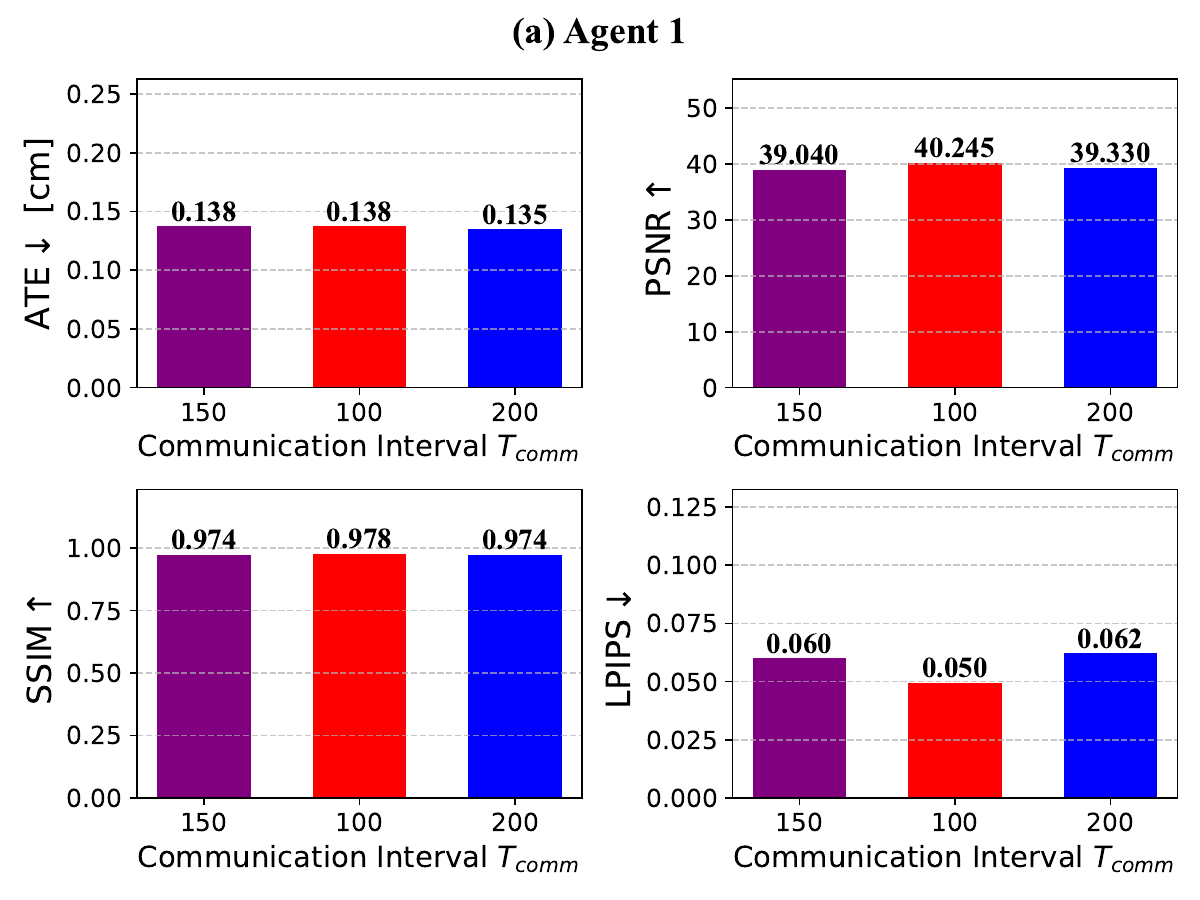}
\includegraphics[width=0.49\textwidth]{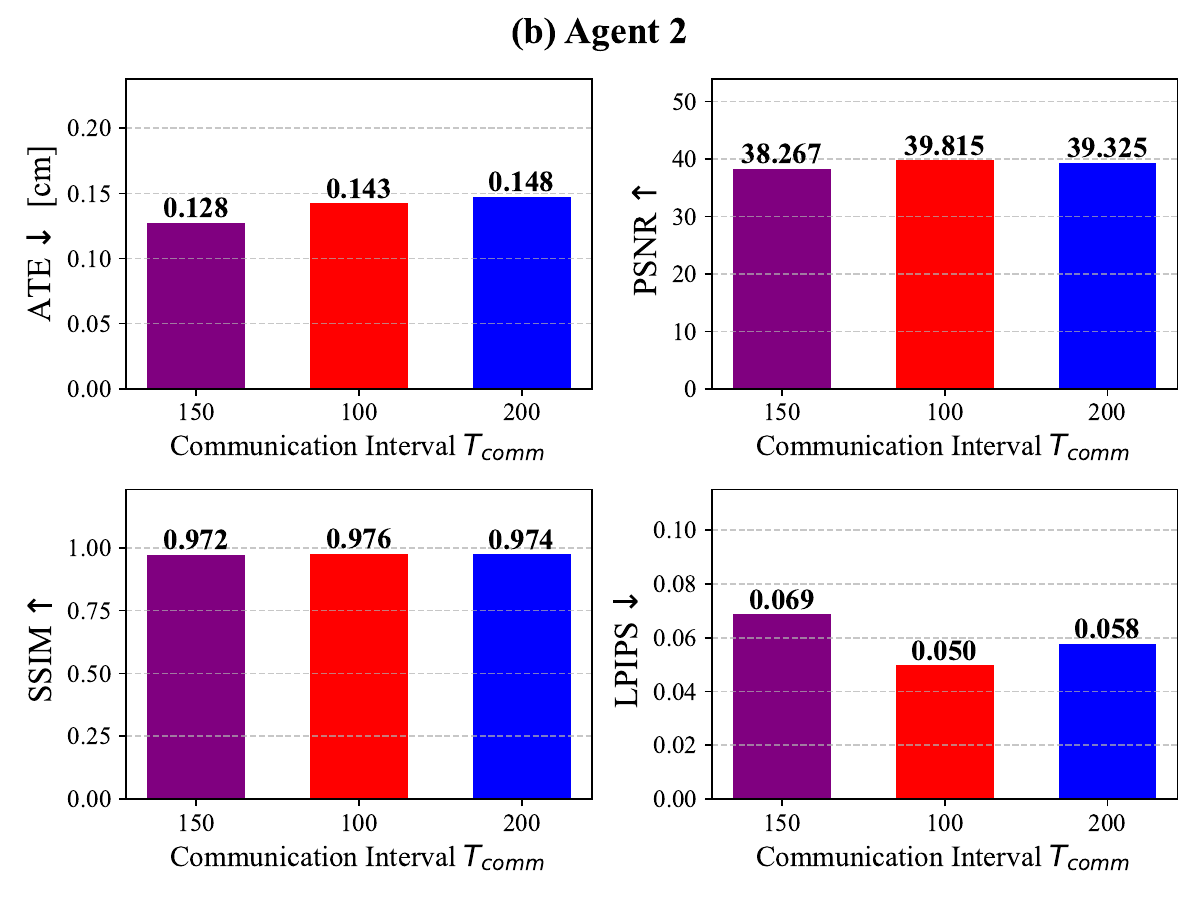}
\includegraphics[width=0.49\textwidth]{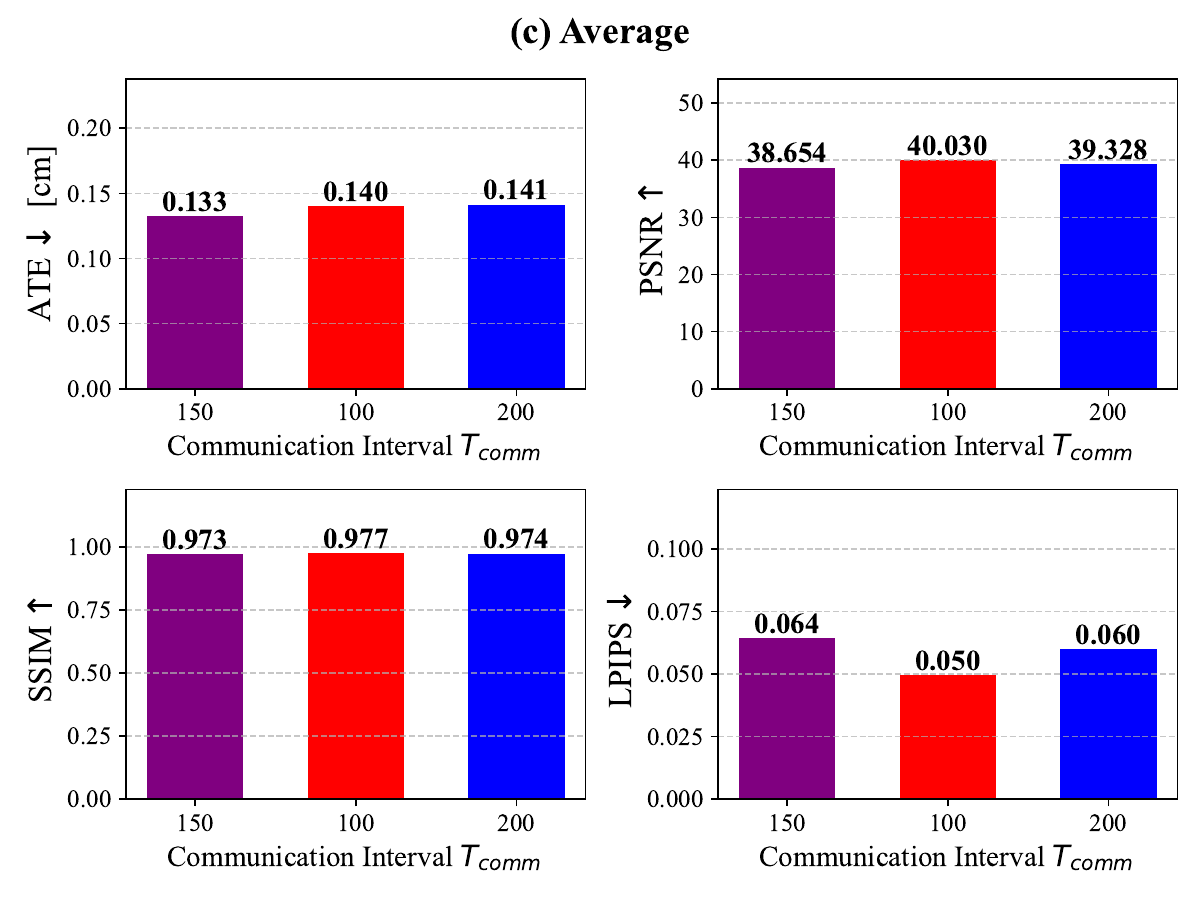}
\caption{\textbf{Hyperparameter sensitivity  analysis of the inter-agent communication interval $T_{Comm}$ on \textit{\textit{\textit{Multi-agent Replica}}}. } } 
\label{fig:sensitivity_comm_multiagent_replica}\vspace{-2mm}
\end{figure}

\begin{figure}[t!]
\centering \setlength{\abovecaptionskip}{0.2cm}
\includegraphics[width=0.49\textwidth]{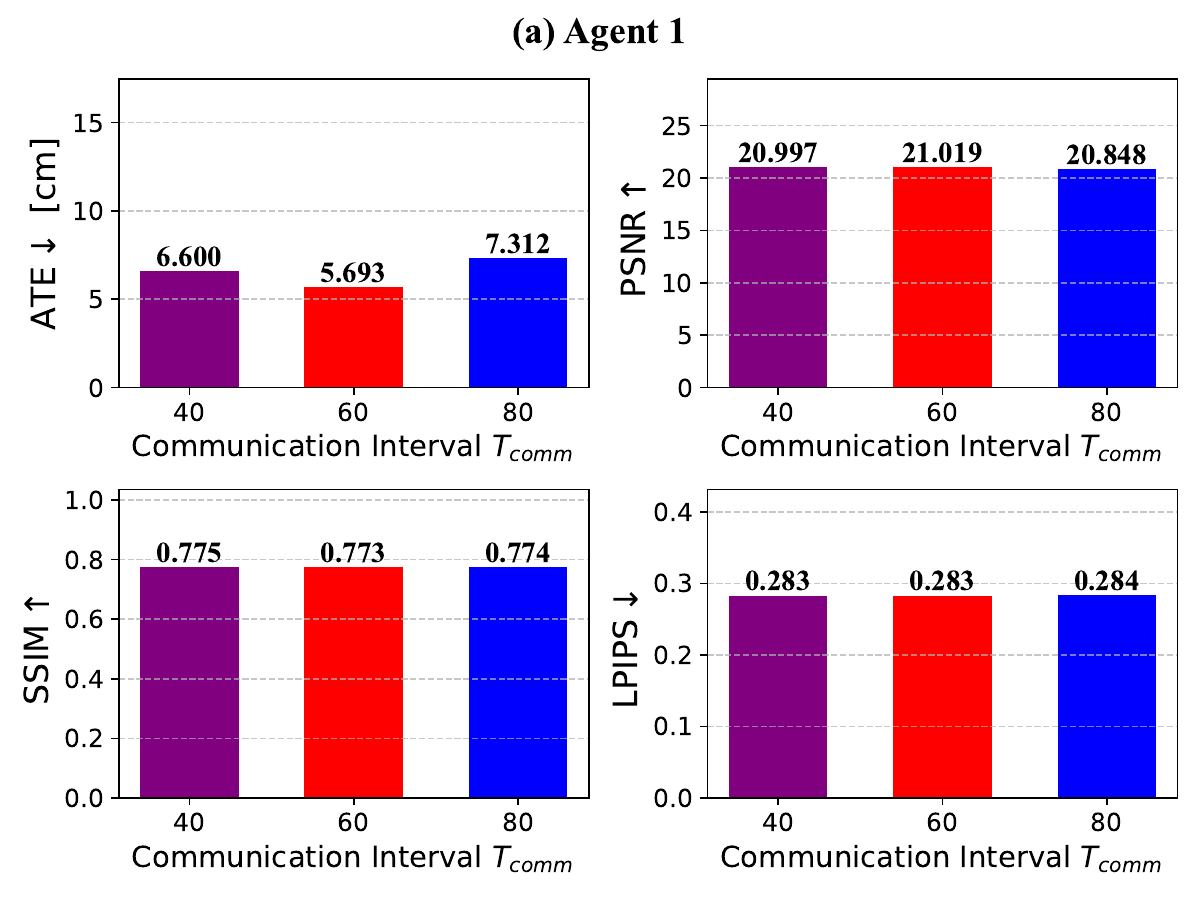}
\includegraphics[width=0.49\textwidth]{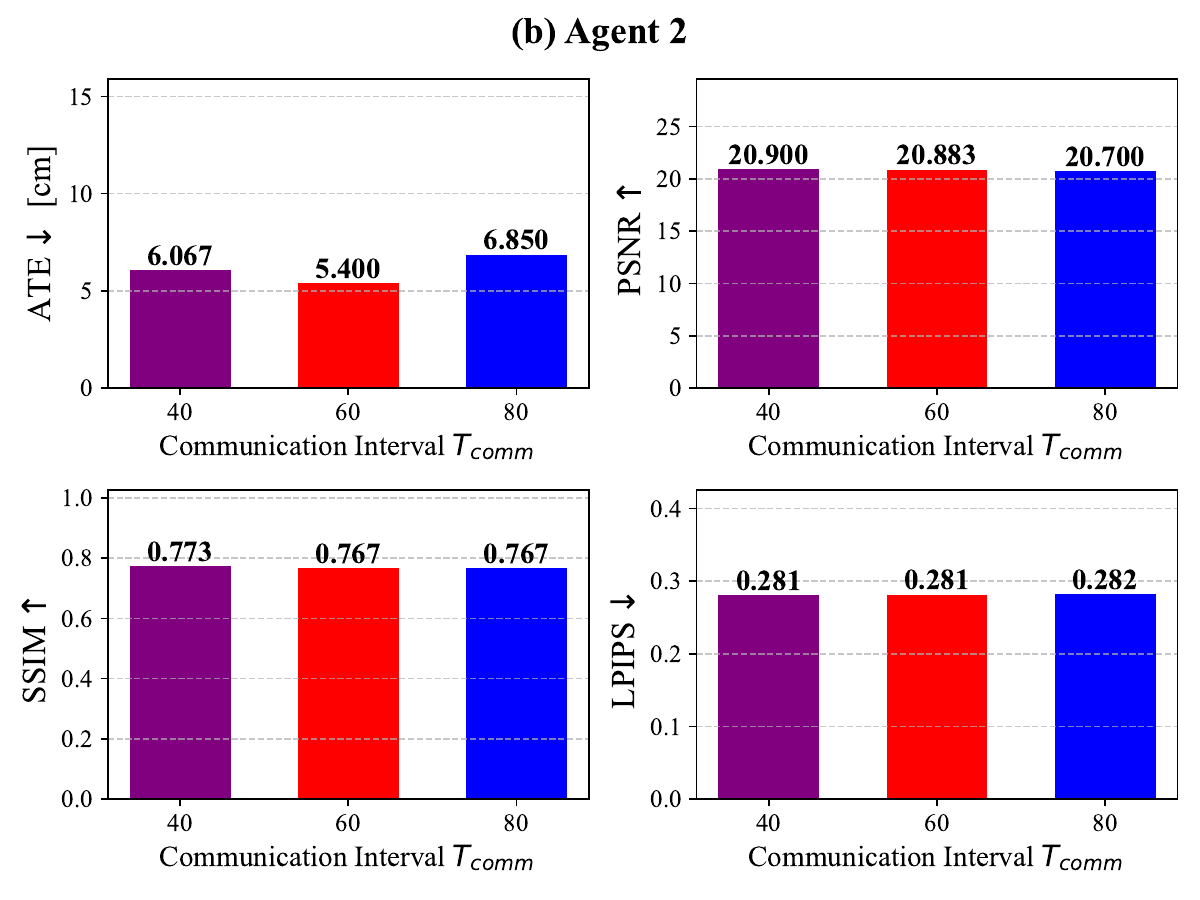}
\includegraphics[width=0.49\textwidth]{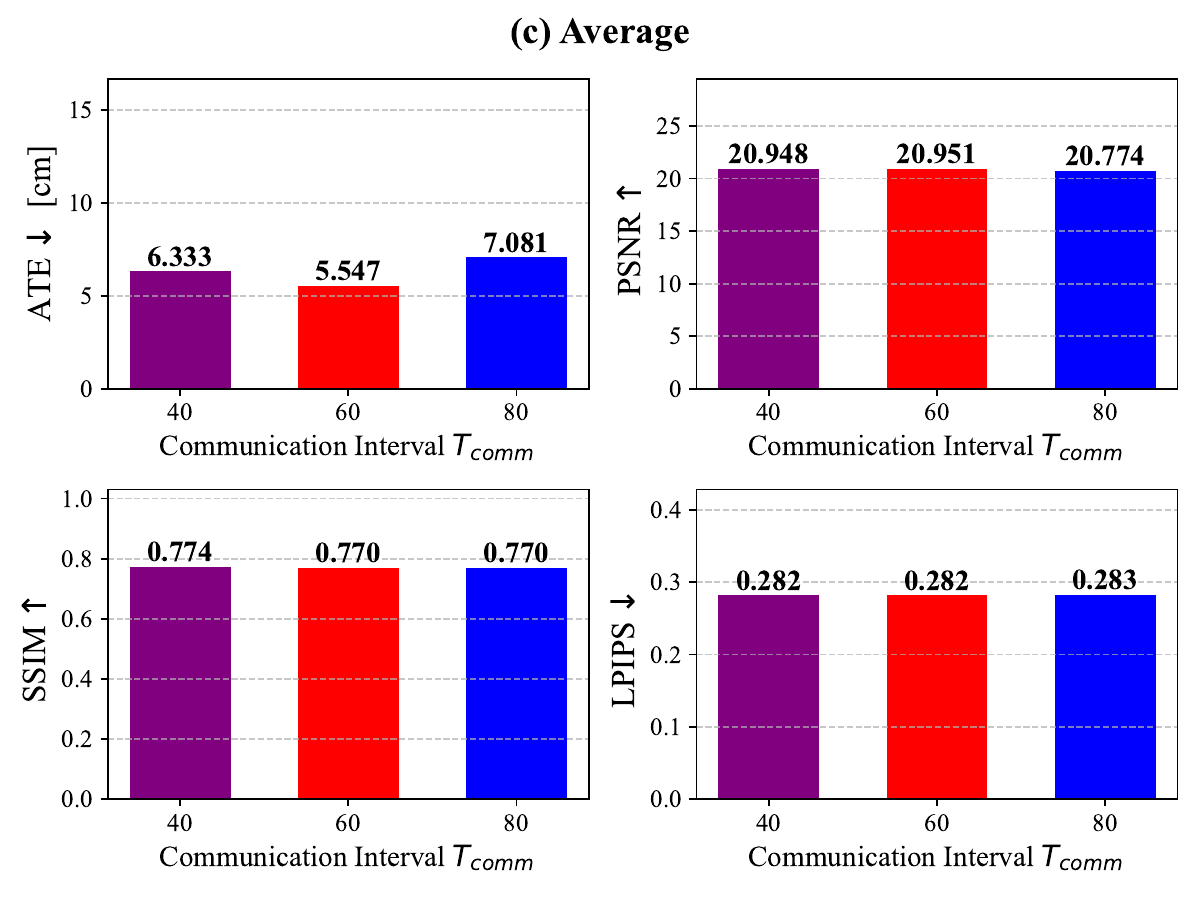}
\caption{\textbf{Hyperparameter sensitivity  analysis of the inter-agent communication interval $T_{Comm}$ on the \textit{\textit{\textit{7Scenes}}} dataset. } } 
\label{fig:sensitivity_comm_multiagent_7scenes}\vspace{-2mm}
\end{figure}

\begin{comment}
\section{Rationale}
\label{sec:rationale}
% 
Having the supplementary compiled together with the main paper means that:
% 
\begin{itemize}
\item The supplementary can back-reference sections of the main paper, for example, we can refer to \cref{sec:intro};
\item The main paper can forward reference sub-sections within the supplementary explicitly (e.g. referring to a particular experiment); 
\item When submitted to arXiv, the supplementary will already included at the end of the paper.
\end{itemize}
% 
To split the supplementary pages from the main paper, you can use \href{https://support.apple.com/en-ca/guide/preview/prvw11793/mac#:~:text=Delete%20a%20page%20from%20a,or%20choose%20Edit%20%3E%20Delete).}{Preview (on macOS)}, \href{https://www.adobe.com/acrobat/how-to/delete-pages-from-pdf.html#:~:text=Choose%20%E2%80%9CTools%E2%80%9D%20%3E%20%E2%80%9COrganize,or%20pages%20from%20the%20file.}{Adobe Acrobat} (on all OSs), as well as \href{https://superuser.com/questions/517986/is-it-possible-to-delete-some-pages-of-a-pdf-document}{command line tools}.
\end{comment}